%% file: main.tex
\documentclass[runningheads]{llncs}

\usepackage{eccv}

\usepackage{eccvabbrv}

\usepackage{graphicx}
\usepackage{booktabs}

\usepackage[accsupp]{axessibility}  

\definecolor{citecolor}{HTML}{c03d3e}
\usepackage{hyperref}

\usepackage{orcidlink}

\usepackage{xspace}
\usepackage[dvipsnames]{xcolor}

\definecolor{ogreen}{HTML}{2E7D32}
\definecolor{mred}{rgb}{0.84,0.15,0.16}
\definecolor{mblue}{HTML}{5496c6}
\definecolor{morange}{HTML}{ffa75b}

\definecolor{gqorange}{HTML}{FA7F6F}

\newcommand{\pskip}{\vspace{0in}}

\newcommand{\ours}[0]{Diffusion Reward\xspace}

\usepackage{graphicx}
\usepackage{amsmath}
\usepackage{amssymb}
\usepackage{booktabs}
\usepackage{amsmath}
\usepackage{amssymb}
\usepackage{comment}
\usepackage{bm}
\usepackage{bbm}
\usepackage{algorithm}
\usepackage{algpseudocode}
\usepackage{mathtools}
\usepackage{multirow}
\usepackage{multicol}

\usepackage{multicol}
\usepackage{multirow}
\usepackage{colortbl}
\usepackage{booktabs}   
\usepackage{bbding} 

\newcommand{\ourcell}{\cellcolor{gray!10}}
\newcommand{\ci}[1]{\tiny{\textcolor{gray}{~($\pm #1$)}}}
\def\arrvline{\hfil\kern\arraycolsep\vline\kern-\arraycolsep\hfilneg}

\usepackage[font=scriptsize,labelfont=bf]{caption}
\usepackage{wrapfig}

\begin{document}

\title{Diffusion Reward: Learning Rewards via \\ Conditional Video Diffusion} 

\titlerunning{Diffusion Reward: Learning Rewards via Conditional Video Diffusion}

\author{Tao Huang\inst{1,2,5,*} \and
Guangqi Jiang\inst{1,3,*} \and
Yanjie Ze\inst{1} \and Huazhe Xu\inst{4,1,5}}

\authorrunning{T. Huang et al.}

\institute{Shanghai Qi Zhi Institute \and The Chinese University of Hong Kong \and
Sichuan University \and Tsinghua University, IIIS \and Shanghai AI Laboratory \\
\href{https://diffusion-reward.github.io/}{\textbf{\color{citecolor}{diffusion-reward.github.io}}} \\
\email{taou.cs13@gmail.com},~ \email{luccachiang@gmail.com.}  
}

\maketitle

\def\thefootnote{*}\footnotetext{Equal contribution to this work.}
\begin{abstract}
Learning rewards from expert videos offers an affordable and effective solution to specify the intended behaviors for reinforcement learning (RL) tasks. 
In this work, we propose \ours, a novel framework that
learns rewards from expert videos via conditional video diffusion models for solving complex visual RL problems. Our key insight is that lower generative diversity is exhibited when conditioning diffusion on expert trajectories. \ours is accordingly formalized by the negative of conditional entropy that encourages productive exploration of expert behaviors. We show the efficacy of our method over robotic manipulation tasks in both simulation platforms and the real world with visual input. Moreover, \ours can even solve unseen tasks successfully and effectively, largely surpassing baseline methods. Project page
and code: \href{https://diffusion-reward.github.io/}{\color{citecolor}{diffusion-reward.github.io}}.
\end{abstract}

\input{sec/1_intro}
\input{sec/2_related}

\input{sec/3_method}

\input{sec/4_exp}

\input{sec/5_conclusion}
\clearpage  

\section*{Acknowledgements}
This work is supported by the National Key R\&D Program of China (2022ZD016
1700). We thank all the reviewers for their insightful comments.
%
%
\bibliographystyle{splncs04}
\bibliography{main}

\input{sec/6_appendix}

\end{document}

%% file: sec/1_intro.tex
\section{Introduction}
\label{sec:intro}
Reward specification is a fundamental challenge in reinforcement learning (RL), determining the effectiveness and alignment of an agent's learned behavior with the intended objectives. Manually designing dense reward functions is both labor-intensive and sometimes infeasible, particularly in real-world tasks such as robotic manipulation~\cite{singh2019end}, where obtaining privileged state information is difficult. As an alternative, using sparse rewards is often favorable in these scenarios because of its low requirement for manual effort~\cite{nasiriany2019planning}. Nonetheless, the insufficient supervision provided by sparse rewards hinders the sample efficiency of RL, potentially imposing an unaffordable burden on data collection.

\begin{figure}[h]
\centering
    \includegraphics[width=0.45\linewidth]{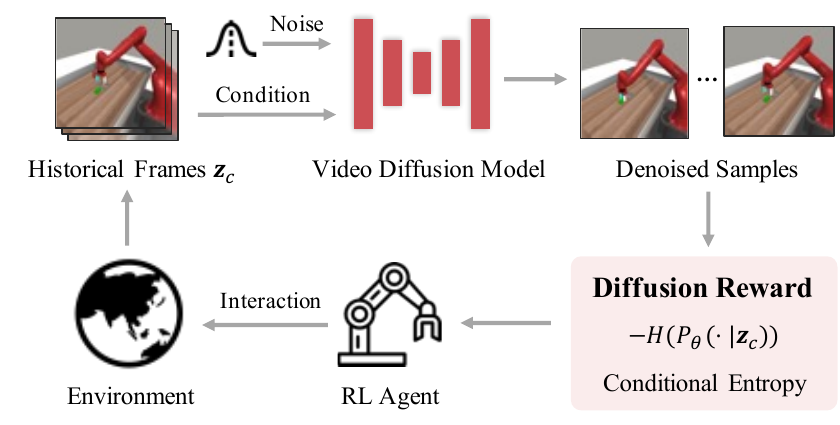}
    \hfill
    \medskip
    \includegraphics[width=0.48\linewidth]{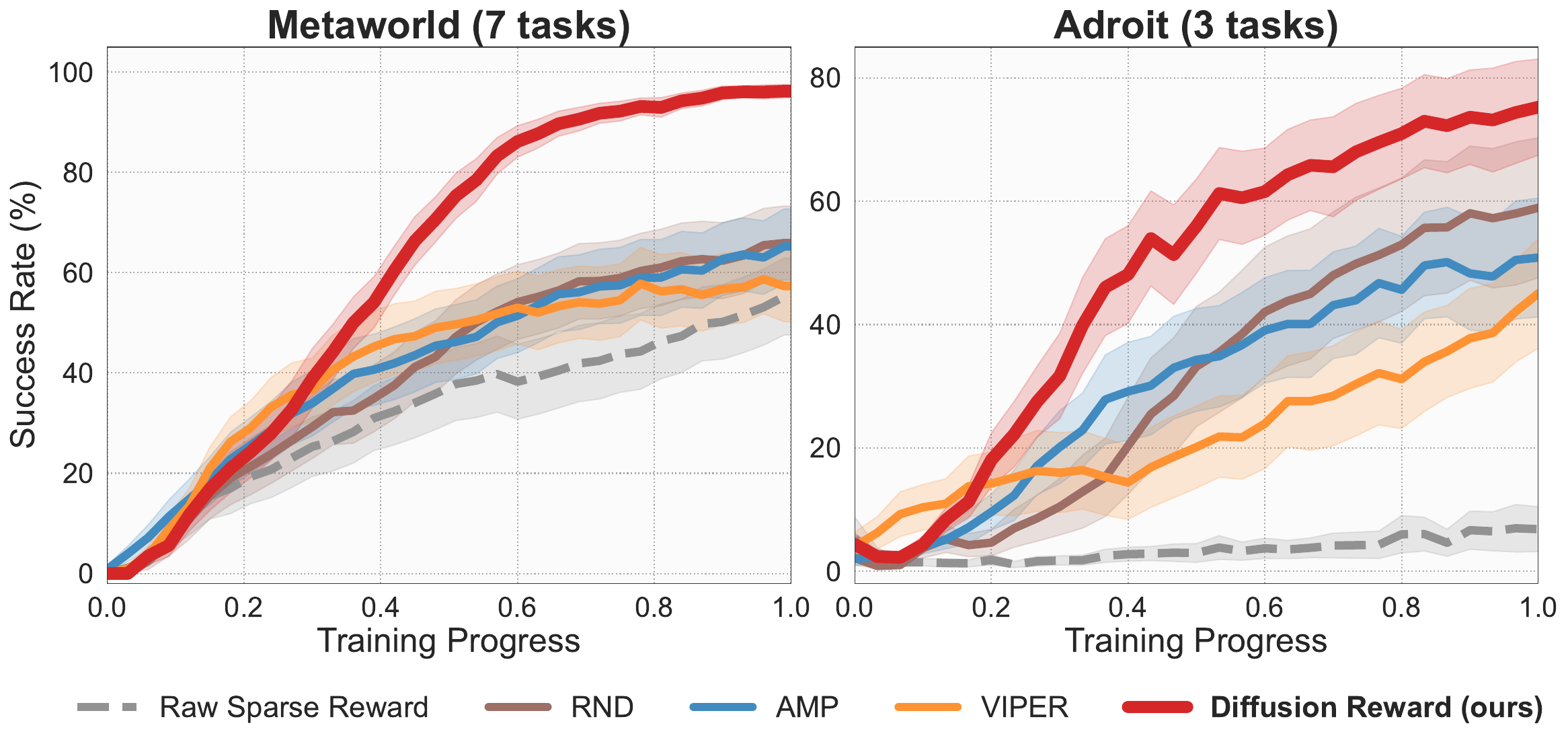}
    \caption{\textbf{Overview of \ours.} (\textit{left})  We present a reward learning framework in RL using video diffusion models pretrained with expert videos. We perform diffusion processes conditioned on historical frames to estimate conditional entropy as rewards to encourage RL exploration of expert-like behaviors. (\textit{right}) The mean success rate of 10 visual robotic manipulation tasks demonstrates the effectiveness of our proposed \ours over 5 runs. Shaded areas are standard errors. }
    \label{fig:overview}
\end{figure}

\begin{figure*}[t]
\begin{minipage}{0.5\linewidth}
    \centering
    \includegraphics[width=1\linewidth]{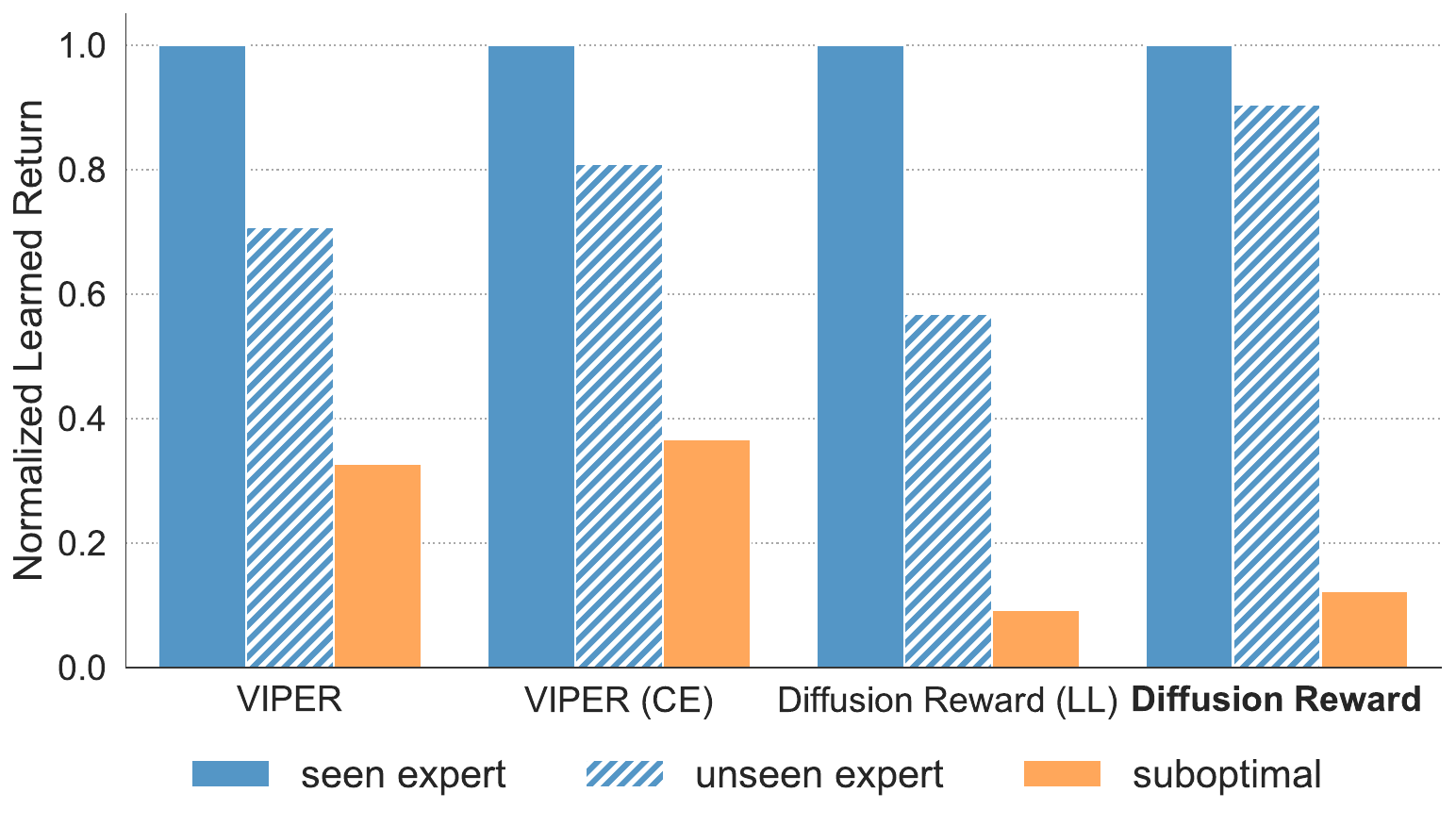}
    \caption{\textbf{Analysis of video models and rewards.} 
    \textit{VIPER (CE)} and \textit{Diffusion Reward (LL)} replace original rewards with conditional entropy (CE) and log-likelihood (LL), respectively. CE-based rewards assign near-optimal rewards to unseen expert videos. Such a boost is enhanced by the strong modeling ability of diffusion models. Results are averaged over 7 MetaWorld tasks. \textit{Suboptimal} represents videos with 25\% randomly-taken actions.}
    \label{fig:e_ll}
\end{minipage}
\hfill
\begin{minipage}{0.45\linewidth}
    \centering
    \includegraphics[width=1\linewidth]{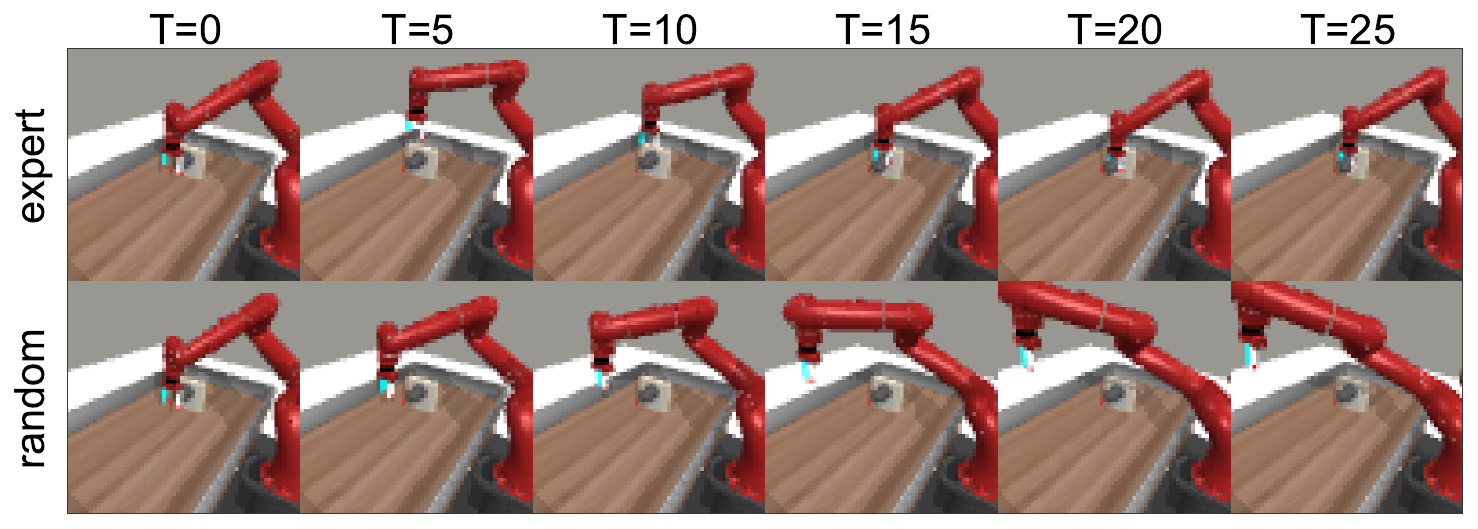}\\%
    \includegraphics[width=1.0\textwidth]{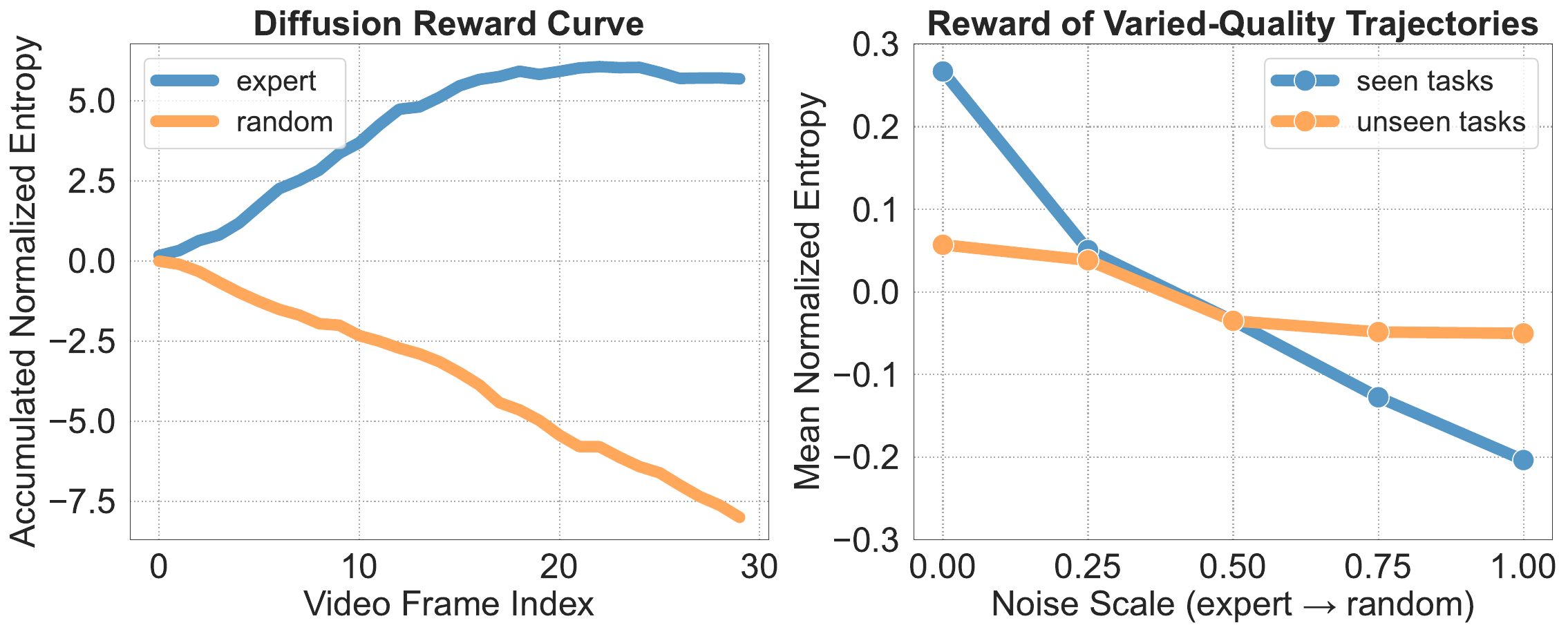}
    \caption{\textbf{Reward analysis.} (\textit{left}) Our learned rewards are aggregately higher for expert behaviors over 7 tasks from MetaWorld. (\textit{right}) Varied-quality trajectories can be distinguished from seen and unseen tasks. (\textit{top}) Examples of trajectories.}
    \label{fig:cond_entropy}
\end{minipage}
\end{figure*}

Learning reward functions from expert videos presents a promising solution because of the low effort of video collection and dense task-execution information contained in the videos~\cite{Chen2021LearningGR,XIRL}. Among these, generative models have been naturally investigated by researchers to extract informative rewards unsupervisedly for RL training~\cite{2021-TOG-AMP,Torabi2018GenerativeAI}. One classical approach builds on generative adversarial learning to learn a discriminative reward that distinguishes between agent and expert behaviors. While straightforward, these methods underutilize the temporal information, whose importance has been shown in solving RL~\cite{Zheng2023TACOTL}, and performance is brittle to the adversarial training. In light of these issues, recent work pre-trains VideoGPT~\cite{VideoGPT} to encode temporal information, and directly use the predicted log-likelihood as rewards~\cite{VIPER}. However, they encounter difficulties in modeling complex expert video distributions and accordingly the proposed rewards, especially in tasks with intricate dynamics. As shown in Figure~\ref{fig:e_ll}, a discernible decline in the learned return is evident for out-of-distribution expert videos, despite their alignment with optimal behaviors in in-distribution videos.

Video diffusion models~\cite{ho2022video} --- which have exhibited remarkable performance in the domain of computer vision --- have shown great power in capturing the complex distribution of videos, such as text-to-video generation~\cite{ho2022imagen, esser2023structure} and video editing~\cite{ceylan2023pix2video, molad2023dreamix}. Recent studies have also investigated their potential in modeling expert videos to serve as generalizable planners in robotic manipulation tasks~\cite{janner2022planning,du2023learning,Ko2023Learning}. Despite these advancements, extracting informative rewards from video diffusion models remains an understudied area, while shining great prospects in guiding RL to acquire expertise from videos and generalize to unseen tasks. 

In this work, we propose \textbf{\ours}, a reward learning framework that leverages video diffusion models to capture expert video distribution and pre-train dense reward function for visual RL. Our key insight is that diffusion conditioned on expert-like trajectories exhibits lower generative diversity than that conditioned on expert-unlike ones. This rationale naturally instructs RL exploration on expert-like behaviors by seeking low generative diversity. We accordingly estimate entropy conditioned on historical frames, which formalizes our insight, and augment it with a novelty-seeking reward~\cite{Burda2018ExplorationBR} and spare environment reward to form dense rewards for efficient RL. In addition, to accelerate the reward inference, we perform latent diffusion processes by utilizing vector-quantized codes~~\cite{VQDiffusion} for compressing high-dimensional observations. 

We empirically validate the effectiveness of our framework through experiments on 10 visual robotic manipulation tasks, including 7 gripper manipulation tasks from MetaWorld~\cite{Yu2019MetaWorldAB} and 3 dexterous manipulation tasks from Adroit~\cite{Rajeswaran2017LearningCD}, exhibiting 38\% and 35\% performance improvements over the best-performing baselines, respectively. Additionally, we surprisingly observe favorable zero-shot generalization performance of our learned reward on unseen tasks. Furthermore, reasonable rewards produced on real robot tasks and associated favorable offline RL performance demonstrate the real-world applicability of Diffusion Reward. Figure~\ref{fig:overview} overviews this work. The primary contributions of this work can be summarized as follows:
\begin{itemize}
    \item We introduce Diffusion Reward, a novel reward learning framework that leverages the generative modeling capabilities of video diffusion models to provide informative dense reward signals for efficient RL.
    \item We demonstrate significant outperformance of our framework compared to baselines across robotic manipulation tasks in simulation and the real world.
    \item We observe that our pretrained reward model can instruct RL on unseen tasks and produce reasonable rewards on real robot tasks. 
\end{itemize}

%% file: sec/2_related.tex
\section{Related Work}
\label{sec:related work}
\paragraph{Learning rewards from videos.} Extracting rewards from expert videos provides an affordable but effective solution for reward specification in RL~\cite{Chen2021LearningGR,sermanet2016unsupervised,Sermanet2017TimeContrastiveNS}. Several studies have endeavored to learn dense rewards that indicate task progress by measuring the distance between current observation and goal image in the latent space~\cite{VIP, XIRL, HOLD}. While promising, obtaining goal images out of simulation can be challenging, thus limiting their applicability to open-world tasks~\cite{Lynch2020LanguageCI}. In contrast, generative models have been widely explored to extract rewards without access to future information. One representative approach, belonging to state-only imitation learning\cite{torabi2018behavioral,radosavovic2021state,zhu2020off}, builds on generative adversarial learning~\cite{Torabi2018GenerativeAI,2021-TOG-AMP} to distinguish expert-like and expert-unlike behaviors with an online paradigm. This idea is further refined in \cite{VIPER}, wherein a reward function is pretrained from videos, and the log-likelihood is predicted as rewards. Nonetheless, it encounters challenges in accurately modeling expert video distributions in complex tasks featuring intricate dynamics, thereby rendering them susceptible to generating unproductive rewards. In contrast, our methods leverage the powerful modeling abilities of diffusion models and estimate the negative of conditional as more discriminative rewards to expedite RL exploration.

\pskip
\paragraph{Diffusion models for RL.} Diffusion models have been widely investigated for RL to, for instance, improve the policy expresiveness~\cite{wang2022diffusion,hansen2023idql,chi2023diffusionpolicy}, and augment experience~\cite{yu2023scaling,chen2023genaug}. Apart from these, some works directly learn the diffusion models from offline data unconditional planners~\cite{janner2022planning}, or conditional planners specified by task returns~\cite{ajay2022conditional}. The idea of conditional diffusion is further investigated in \cite{du2023learning,Ko2023Learning} with text as task specification, where video diffusion models serve as planners associated with inverse dynamics. Unlike these methods, we intend to learn informative rewards via video diffusion models conditioned on historical frames to accelerate online RL. Such historical conditioning has been used in \cite{hu2023instructed} to inform trajectory generation, which differs from our focus on reward learning as well. Our work is also close to \cite{ExtractRF}, which extracts rewards from two diffusion models that fit different behaviors in state-based tasks. Differently, we pretrain reward models without action labels and additional task-agnostic data. 

\pskip
\paragraph{Pretraining for RL.} Numerous methods have been explored for directly pretraining effective policies in RL from data. These methods include those employing offline RL techniques before online finetuning~\cite{nair2020awac,lee2022offline,kun2023uni} as well as approaches learning behavioral (skill) priors~\cite{pertsch2021accelerating, singh2020parrot, rao2021learning}. However, they require the additional collection of rewards, action labels, or task-agnostic data, which poses certain sacrifices. In contrast, our approach focuses on pretraining rewards solely from action-free expert-only videos, alleviating the need for such additional data sources. Some studies have also explored pretraining visual representations using datasets similar to ours~\cite{xiao2022masked,nair2023r3m,parisi2022unsurprising,yuan2022pre}. Nevertheless, these approaches do not extract dense task-execution signals to guide RL.

%% file: sec/3_method.tex
\section{Preliminaries}

\begin{table}[t]
    \begin{minipage}{0.56\linewidth}
	\centering
	\small
	\renewcommand{\tabcolsep}{2.5pt}
\resizebox{0.98\columnwidth}{!}{
	\begin{tabular}{@{}ccccc}
            \toprule
		Data & Model & SSIM ($\uparrow$) & PSNR ($\uparrow$) & LPIPS ($\downarrow$)
		\\\midrule
           \multirow{2}{*}{Real} & VideoGPT &0.768\ci{0.141} & 23.35\ci{4.13} &0.1017\ci{0.0670} \\\noalign{\vskip 0.1ex}
           & VQ-Diffusion &\ourcell\textbf{0.856}\ci{0.114} & \ourcell\textbf{27.46}\ci{5.24} &\ourcell\textbf{0.0596}\ci{0.0450} \\
           \midrule
		 \multirow{2}{*}{Sim.} & VideoGPT &0.967\ci{0.021} & 32.28\ci{3.15} &0.0067\ci{0.0032}  \\ 
         &VQ-Diffusion &\ourcell\textbf{0.976}\ci{0.009} &\ourcell\textbf{33.55}\ci{2.31} &\ourcell\textbf{0.0053}\ci{0.0029}  \\ 
		\bottomrule
	\end{tabular}}	
  \caption{\textbf{Quantitative comparison of video models.} Results are evaluated on real dexterous robotic manipulation videos and simulation videos from 7 MetaWorld tasks, demonstrating that the video diffusion model generates videos of higher quality than VideoGPT.}
     \label{tab:quant-video-prediction}
    \end{minipage}
\hfill
    \begin{minipage}{0.42\linewidth}
	\centering
	\small
	\renewcommand{\tabcolsep}{2.5pt}
\resizebox{0.98\columnwidth}{!}{
	\begin{tabular}{@{}lccc}
            \toprule
    \multirow{2}{*}{$\epsilon$-greedy traj.}  & \multicolumn{3}{c}{\textbf{Distances as Diversity Metrics}}  \\ \cmidrule(lr){2-4}
		 & SSIM ($\uparrow$) & PSNR ($\uparrow$) & LPIPS ($\downarrow$)
		\\\midrule
		 0\% (expert) &\ourcell\textbf{0.941}\ci{0.054} & \ourcell\textbf{29.91}\ci{5.52} &\ourcell\textbf{0.0171}\ci{0.0166}  \\ 
            25\% &0.901\ci{0.063} &26.01\ci{4.62} & 0.0371\ci{0.0304}  \\ 
            50\% &0.894\ci{0.065} &25.65\ci{4.29} & 0.0421\ci{0.0297}  \\ 
            75\% &0.844\ci{0.066} &22.60\ci{2.53} & 0.0645\ci{0.0370}  \\ 
            100\% (rand.) & 0.829\ci{0.730} & 22.25\ci{3.43} & 0.0597\ci{0.0364}  \\ 
		\bottomrule
	\end{tabular}}
  \caption{\textbf{Generative diversities.} Expert trajectories over 7 MetaWorld tasks, seen during training, show the lowest generative diversity, which is proportional to distances among generated videos \cite{zhu2017toward}.}
     \label{tab:diversity}
    \end{minipage}
\end{table}

\paragraph{Problem formulation.} We consider an RL agent that interacts with the environment modeled as a finite-horizon Markov Decision Process (MDP), which is defined by a tuple $\mathcal{M}=\langle \mathcal{S},\mathcal{A}, \mathcal{T}, \mathcal{R}, \gamma\rangle$, where $\mathcal{S}$ is the state space, $\mathcal{A}$ is the action space, $\mathcal{T}$ is the environment transition function, $\mathcal{R}$ is the reward function, and $\gamma$ is the discount factor. The goal of the RL agent is to learn an optimal policy $\pi$ that maximizes the expected return $\mathbb{E}[\sum_{k=0}^{K-1} \gamma^k r_k]$ over $K$ timesteps.

In this work, we focus more specifically on high-dimensional state space with a binary sparse (i.e., 0/1 reward as a success indicator of the task). Particularly, we consider RGB images $\bm{x}\in\mathbb{R}^{H\times W \times 3}$ as the state observed by the agent. This setting is motivated by the real-world application of RL such as robotics, where vision-based sensory data is more available and specifying sophisticated reward often requires tedious hand-engineering, sometimes even infeasible. However, this poses great difficulty to an RL agent, as a large amount of interaction with environments is often required, known as sample efficiency.

\pskip
\paragraph{Expert videos.} To improve the sample efficiency of RL, we assume a set of unlabeled videos generated by the expert policies are accessible by the agent. Notably, these videos are action-free and gathered from multiple tasks without task identification. We denote it as $\mathcal{D}=\{\mathcal{D}^{1}, \mathcal{D}^{2},..., \mathcal{D}^{N}\}$, where $\mathcal{D}^{i}$ is the demonstrated video set of task $i$. Each video set $\mathcal{D}^i$ consists of multiple expert videos $\tau=\{\bm{x}_0, \bm{x}_{1},...,\bm{x}_{K-1}\}\in \mathcal{S}^K$. Our goal is to pretrain effective reward functions from expert videos to accelerate the online exploration of RL. We present our reward learning framework in the following section. 


\section{Method}

\begin{figure*}[t]
\resizebox{0.55\columnwidth}{!}{
\begin{minipage}{0.7\linewidth}
\begin{algorithm}[H]
	\caption{\ours for online RL} \label{alg:inference}
	\begin{algorithmic}[1]
            \Statex \textcolor{gray}{// Pretrain reward model from expert videos}
            \State Collect expert videos $\mathcal{D}$ from $K$ tasks 
		\State Train diffusion model $p_\theta$ on expert videos $\mathcal{D}$
            \Statex \textcolor{gray}{// Downstream RL with learned rewards}
		\While{not converged}
            \State Act $a_k\sim\pi(\cdot|\bm{x}_k)$ 
            \State Generate $M$ denoised samples $\tilde{\bm{z}}_{k}^{0:T}\sim p_\theta(\bm{z}_k^{0:T}|\bm{z}_c)$ 
            \State Estimate entropy $H(p_\theta(\cdot|\bm{z}_{c}))$ with $\tilde{\bm{z}}_{k}^{0:T}$ \Comment{\textcolor{gray}{Eq.~\eqref{eq:estimation}}}
            \State Compute Diffusion Reward $r_k \leftarrow r^{\mathrm{diff}}_k$ \Comment{\textcolor{gray}{Eq.~\eqref{eq:final_reward}}}
            \State Step environment $\bm{x}_{k+1}\sim \mathcal{T}(\bm{x}_k, a_k)$
            \State Store transition $(\bm{x}_k, a_k,r_k,\bm{x}_{k+1})$
            \State Update policy $\pi$ and $r^{\mathrm{rnd}}$ with RL algorithm 
		\EndWhile
	\end{algorithmic}
        \end{algorithm}
\end{minipage}}
\hfill
\begin{minipage}{0.4\linewidth}
  \centering
    \includegraphics[width=0.9\linewidth]{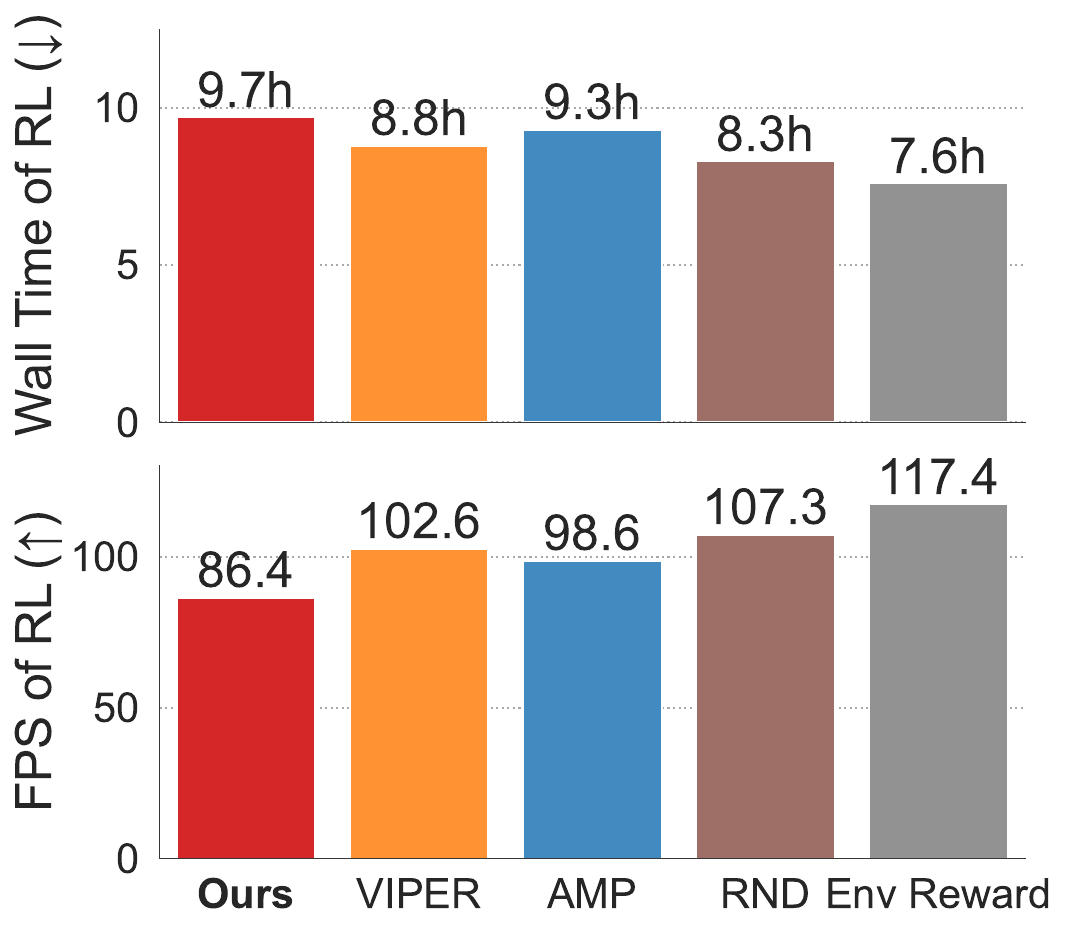}
    \caption{\textbf{Time efficiency} on Door task with a NVIDIA A40 GPU.} 
    \label{fig:time}
\end{minipage}
\end{figure*}

We introduce \ours, a novel framework that learns rewards from expert videos via conditional video diffusion for solving downstream visual RL problems, as illustrated in Figure~\ref{fig:overview} and Algorithm~\ref{alg:inference}. 
In Section~\ref{subsec:ldm}, we first model the expert videos with the video diffusion model in the latent space. In Section~\ref{subsec:rew}, we formalize our key insight of \ours by estimating the history-conditioned entropy and using its negative as rewards for RL. 

\subsection{Expert Video Modeling via Diffusion Model}\label{subsec:ldm}
Diffusion models~\cite{sohl2015deep} are probabilistic models that aim to model the data distribution by gradually denoising an initial distribution through a reverse diffusion process~\cite{ho2020denoising}. These models showcase their power in capturing highly complex distributions and generating samples that exhibit intricate dynamics, motion, and behaviors in RL literature~\cite{janner2022planning, ajay2022conditional}. Unlike prior works that model expert videos as planners, we aim to learn reward functions from the diffusion model trained on expert videos for RL. This motivates our video models to achieve fast inference speed and encode sufficient temporal information.

\pskip
\paragraph{Latent diffusion process.} Specifically, we first train an encoder unsupervisedly from expert videos to compress the high-dimensional observations. Here we use the VQ-GAN method~\cite{esser2021taming} to represent the image $\bm{x}$ with a vector-quantized latent code $\bm{z}=Q(E(\bm{x}))$, where $E$ is the encoder and $Q$ is the element-wise quantizer. The whole video is then represented by a sequence of latent variables $\tau=\{\bm{z}_0, \bm{z}_{1},...,\bm{z}_{K-1}\}$, where we overwrite the definition of $\tau$ without ambiguity. Subsequently, the forward process applies noise $\epsilon$ in the latent space at each time step $t\in 0, ...,T$ to the data distribution $\bm{z}_k$, resulting in a noisy sample $\bm{z}_{k}^t$, where $\bm{z}_{k}^t=\sqrt{\bar{\alpha}_t}\bm{z}_{k}^0+\sqrt{1-\bar{\alpha}_t}\bm{\epsilon} $, and $\bar{\alpha}$ is the accumulation of the noise schedule over past timesteps. To fit data distribution, we learn a parameterized variant of noise predictor $\bm{\epsilon}_{\theta}(\bm{z}_{k}^t)$ that aims to predict the noise $\bm{\epsilon}$ during the forward process. Then the parameterized reverse process $p_\theta(\bm{z}_{k}^{t-1}|\bm{z}_{k}^{t})$ can be approximated and performed by iterative denoising an initial distribution.

\pskip
\paragraph{Historical frames as condition.} To utilize the temporal information from expert videos with the power of video diffusion, we further condition the reverse process with historical frames, i.e., $p_\theta(\bm{z}_{k}^{t-1}|\bm{z}_{k}^t, \bm{z}_c)$, where $\bm{z}_c$ is the concatenation of $\ell$ historical frames $[\bm{z}_{k-\ell},...,\bm{z}_{k-1}]$ to ensure higher computation efficiency while maintaining enough historical information. This can also be viewed as matching the distribution of expert and agent trajectories~\cite{VIPER}. 
Subsequently, one can perform the reverse process from a randomly sampled noise to generate the latent code of future frames and decode the code for video prediction. In this work, we use VQ-Diffusion~\cite{VQDiffusion} as our choice of video diffusion model due to its good performance and compatibility with vector-quantized latent code, but our framework can in principle adopt any off-the-shelf video diffusion models. Specifically, we first tokenize each latent code $\bm{z}$ indexed by its indices that specify the respective entry in the learned codebook of VQ-GAN, and take as a condition embedding the concatenated tokens. The embedding is then fed to the decoder that contains multiple transformer blocks and softmax layers with cross attention, providing context information for video diffusion.

\subsection{Conditional Entropy as Rewards}\label{subsec:rew}


While previous studies have explored the use of log-likelihood as rewards with video prediction models, exemplified by works such as VIPER \cite{VIPER}, this approach encounters two primary challenges. Firstly, it has difficulty in accurately modeling the distribution of complex expert videos featuring intricate dynamics, as shown in Table~\ref{tab:quant-video-prediction}. Secondly, the moderate video modeling ability leads to undesired rewards. This issue is evident in Figure~\ref{fig:e_ll}, where a significant drop in learned rewards between in-distribution expert videos and out-of-distribution ones, though both sets of videos demonstrate optimal behaviors.

\pskip
\paragraph{Key insight behind Diffusion Reward.} We address these challenges by harnessing the great generative capability of video diffusion models. Our observations indicate increased generation diversity with unseen historical observations (rand.) and reduced diversity with seen ones (expert), as shown in Table~\ref{tab:diversity}. This gives rise to the key insight of our proposed \ours: diffusion conditioned on expert-like trajectories exhibits lower diversity where the agent ought to be rewarded more, and the opposite holds on unexpert-unlike ones. Such diversity discrimination incentivizes RL agents to chase expert-like behaviors.

\pskip  
\paragraph{Estimation of conditional entropy.} To formalize this idea, we seek to estimate the negative conditional entropy given historical frames $\bm{z}_c$, which in principle captures the conditional generation diversity:
\begin{equation}
\label{eq:ce}
    -H(p_\theta(\cdot|\bm{z}_{c}))=\mathbb{E}_{p_\theta(\cdot|\bm{z}_{c})}\left[ \log p_\theta(\bm{z}_{k}|\bm{z}_{c}) \right].
\end{equation} 
One primary challenge is the computation of the entropy in Eq.~\eqref{eq:ce}, which is intractable as we have no explicit form of the conditional distribution~\cite{song2020score}. Therefore, we instead estimate the variational bound of the entropy. Specifically, we first present the variational bound of conditional log-likelihood as follows:
\begin{equation}\label{eq:ll}
\begin{split}
    \log p_\theta(\bm{z}_{k}^0|\bm{z}_{c}) \geq &  \mathbb{E}_{q(\bm{z}_{k}^{0:T}|\bm{z}_c)}\left[\log{\frac{ p_\theta(\bm{z}_{k}^{0:T})} 
    { q(\bm{z}_{k}^{1:T}|\bm{z}_{k}^{0}, \bm{z}_{c})}}\right],
\end{split}
\end{equation}
where $\bm{z}_k^0$ is the denoised prediction of current observation $\bm{z}_k$. This bound can be estimated via noise predictor $\bm{\epsilon}_\theta$~\cite{ho2020denoising,song2020denoising}, or with the closed-form distribution~\cite{sohl2015deep,kingma2013auto} (e.g., discrete multivariate distribution). We use the latter one as our choice of estimation because of its better compatibility with VQ-Diffusion. 

Next, to estimate the whole entropy, we denoise from randomly sampled noise and generate the latent variables $\tilde{\bm{z}}_{k}^{0:T}\sim p_\theta(\bm{z}_k^{0:T}|\bm{z}_c)$, repeating with $M$ times. Subsequently, we use the generated samples from randomized noise $\tilde{\bm{z}}_k^T$ (e.g., random tokens) to estimate the whole conditional entropy term as follows:
\begin{equation}\label{eq:estimation}
\begin{split}
    r^{\mathrm{ce}}(\bm{x}_{k-1}) = \frac{1}{M}\sum_{j=1}^M \log{\frac{ p_\theta(\tilde{\bm{z}}_{k}^{0:T})} 
    { q(\tilde{\bm{z}}_{k}^{1:T}|\tilde{\bm{z}}_{k}^{0}, \bm{z}_{c})}},
\end{split}
\end{equation}

We visualize the aggregated entropy-based reward in Figure~\ref{fig:cond_entropy} and present curves of each task in the Appendix. The results show that conditional entropy can successfully capture the varied generative diversity on different videos, echoed with the aforementioned insight of \ours. 
Notably, here we use the standardized entropy reward $\bar{r}^{\mathrm{ce}}$ to mitigate the burden of hyperparameter tuning, as we observe that the scale of conditional entropy varies significantly across different tasks and domains, partially attributed to the varied objects and environment dynamics. Concretely, the conditional entropy is standardized by the empirical mean and standard deviation of the expert videos: 
\begin{equation}
    \bar{r}^{\mathrm{ce}} = (r^{\mathrm{ce}} - \mathrm{mean}(\mathcal{D}, r^{\mathrm{ce}}))\; /\; \mathrm{std}(\mathcal{D}, r^{\mathrm{ce}}).
\end{equation}

\paragraph{Exploration reward.} As the reward $\bar{r}^{\mathrm{ce}}$ incentivizes the agent to mimic the behavioral patterns of the expert, the exploration may still be prohibitively challenging in complex tasks with high-dimensional input~\cite{VIPER,adeniji2023language}. To alleviate this issue, we incorporate RND~\cite{Burda2018ExplorationBR} as the exploration reward, termed as $r^{\mathrm{rnd}}$.

\pskip
\paragraph{Diffusion Reward.} Lastly, we combine our proposed diffusion-based entropy reward with the exploration reward and the raw sparse task reward $r^{\mathrm{spar}}$ from the environment, yielding the following composite reward function: 
\begin{equation}\label{eq:final_reward}
    r^{\mathrm{diff}} = (1- \alpha)\cdot \bar{r}^{\mathrm{ce}} + \alpha \cdot r^{\mathrm{rnd}} + \beta\cdot r^{\mathrm{spar}},
\end{equation}
where $\alpha$ and $\beta$ are the reward coefficients that steer the effect of each reward. The inclusion of $r^{\mathrm{spar}}$ is crucial, as the complete absence of environmental supervision can impede progress in tackling complex tasks. Furthermore, we opt for a single coefficient $\alpha$ to balance $\bar{r}^{\mathrm{ce}}$ and $r^{\mathrm{rnd}}$, thereby alleviating the burden of hyperparameter tuning. Meanwhile, the coefficient of task reward $\beta$ is set as 1 for all tasks. We present negligible empirical distinctions between this formulation and the one featuring an additional coefficient, and effect of $\beta$ in the Appendix. 


\subsection{Training Details}

We first provide the expert videos from various tasks as the whole dataset for pertaining reward models, which can be generated by scripted policies or other means. Then, we first use VQ-GAN~\cite{esser2021taming} to train the encoder, associated with $8\times 8$ size of codebook across all domains, with an additional perceptual loss~\cite{zhang2018unreasonable} calculated by a discriminator to increase perceptual quality. We subsequently use VQ-Diffusion~\cite{VQDiffusion} for training the conditional video diffusion model, where the number of historical frames $\ell$ is set as 2 for all tasks. For reward inference, the reward coefficient is set as $0.95$ for all tasks except $0$ for the Pen task. We decrease $\alpha$ for the Pen task because a high exploration reward can misguide the robot hand to drop the pen, while a lower one encourages in-hand manipulation. Moreover, during the diffusion process, we also use a DDIM-like sampling strategy~\cite{song2020denoising} to accelerate the diffusion process with denoise steps set as $10$ and repetition of diffusion $M$ as $1$ for all tasks, which shows fair performance and retains high inference speed in Figure~\ref{fig:time}. More details on hyperparameters, network architectures, and downstream RL can be found in the Appendix.

%% file: sec/4_exp.tex
\section{Experiments}


\subsection{Experimental Setup}\label{subsec:setup}

\begin{figure*}[t]
\begin{minipage}{0.45\linewidth}
     \centering
     \begin{subfigure}[b]{0.48\linewidth}
         \centering
         \includegraphics[width=\textwidth]{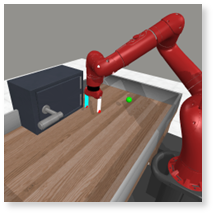}
         \caption{MetaWorld (7 tasks)}
         \label{fig:metaworld}
     \end{subfigure}
     \hfill
     \begin{subfigure}[b]{0.48\linewidth}
         \centering
         \includegraphics[width=\textwidth]{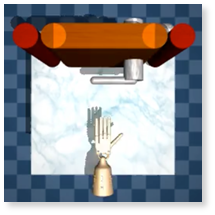}
         \caption{Adroit (3 tasks)}
         \label{fig:adroit}
     \end{subfigure}
     \hfill
        \caption{\textbf{Task visualization.} We evaluate methods on 10 challenging visual RL tasks from MetaWorld and Adroit with visual input and sparse rewards. Tasks are chosen to cover a wide range of manipulation skills.}
        \label{fig:task description}
\end{minipage}
\hfill
\begin{minipage}{0.525\linewidth}
    \centering
    \includegraphics[width=0.9\linewidth]{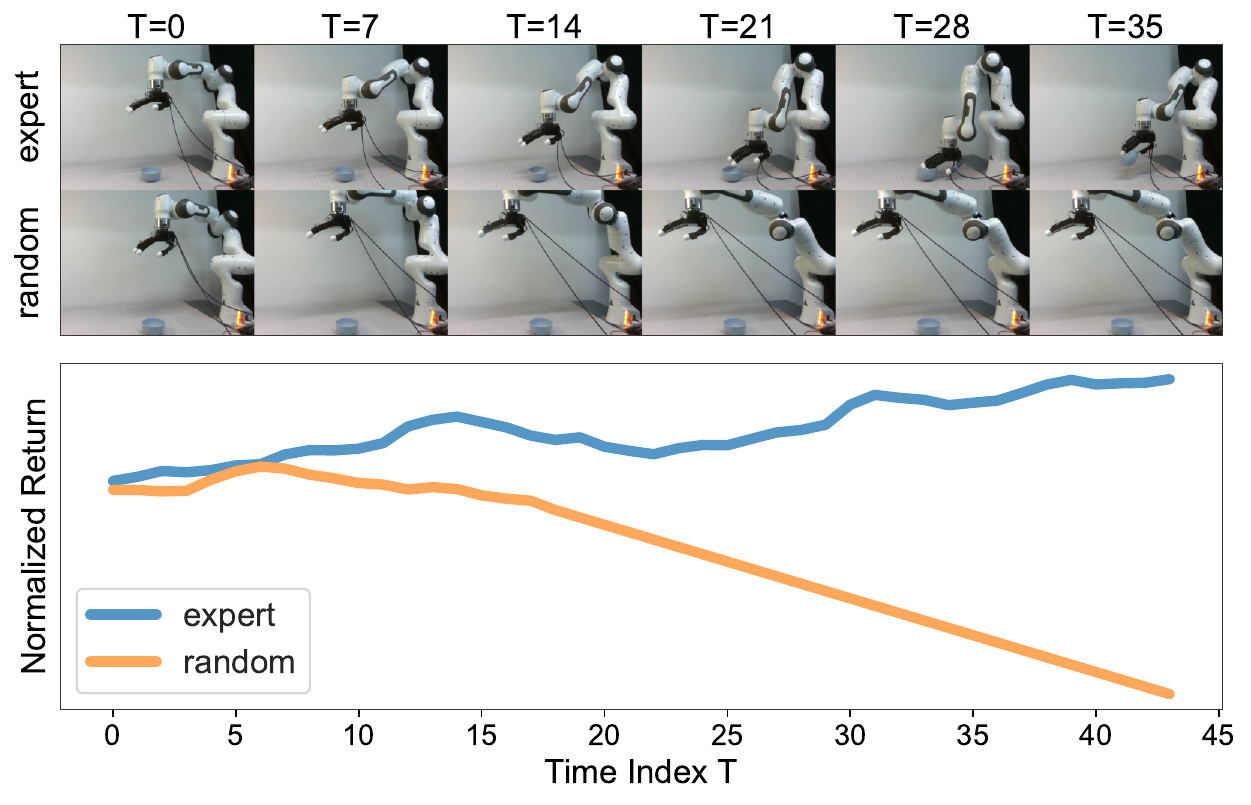}
    \caption{\textbf{Reward curve of real robot videos.} Our method assigns higher rewards to expert videos than to random ones.}
    \label{fig:reward-curve-real-robot}
\end{minipage}
\end{figure*}


\paragraph{Simulation environments.} 
We intend to demonstrate the effectiveness of \ours on 10 complex visual robotic manipulation tasks, including 7 gripper manipulation tasks from MetaWorld~\cite{Yu2019MetaWorldAB} and 3 dexterous hand manipulation tasks from Adroit~\cite{Rajeswaran2017LearningCD}, as visualized in Figure~\ref{fig:task description}. We choose these two simulation environments because of their task diversity and complexity. Each task is associated with $64\times 64$-dimensional RGB images, $\pm 4$ pixel shift augmentation~\cite{yarats2021mastering}, and 0/1 sparse task reward. For pertaining reward models, we collect 20 expert videos for each Metaword task (5039 frames in total) via the scripted policy provided by the official repository, and 50 for Adroit (3346 frames in total) via the policies trained with performant RL method~\cite{VRL3}, while our method has emerging low demand for the data scale (see the Appendix). For downstream RL training, the interaction budget of the Adroit task is set as $3$ million to ensure convergence of using our rewards, and is respectively set for the MetaWorld task due to the noticeably varied task complexity.

\pskip
\paragraph{Baselines.} We compare our method against the followings: 
\begin{itemize}
    \item \textbf{Raw Sparse Reward} that uses the sparse task reward. This comparison tests the benefit of adding learned rewards from videos.
    \item \textbf{Random Network Distillation} (RND,~\cite{Burda2018ExplorationBR}) that encourages exploration with a novelty-seeking reward only. This comparison tests the benefit of rewarding the agent with our pretrained reward.
    \item \textbf{Adversarial Motion Priors} (AMP,~\cite{2021-TOG-AMP}) that online learns a discriminator to discern agent behaviors and expert behaviors based on current observations. This comparison tests the benefit of encoding temporal information in learned reward and utilization of novelty-seeking reward.
    \item \textbf{Video Prediction Rewards} (VIPER,~\cite{VIPER}) and its standardized variant (VIPER-std) that use VideoGPT~\cite{VideoGPT} as video prediction model and predicted log-likelihood of agent observation as reward. This comparison tests the benefit of utilizing the generative capability of video diffusion models and the conditional entropy as a more explorative reward.
\end{itemize}
For a fair comparison, all methods are augmented with task rewards, which generally improve their performance. Besides, we use DrQv2~\cite{yarats2021mastering} as the RL backbone and maintain all settings except reward-pretraining (if exist) identical.

\begin{figure}[!t]
  \centering
  \includegraphics[width=1\linewidth]{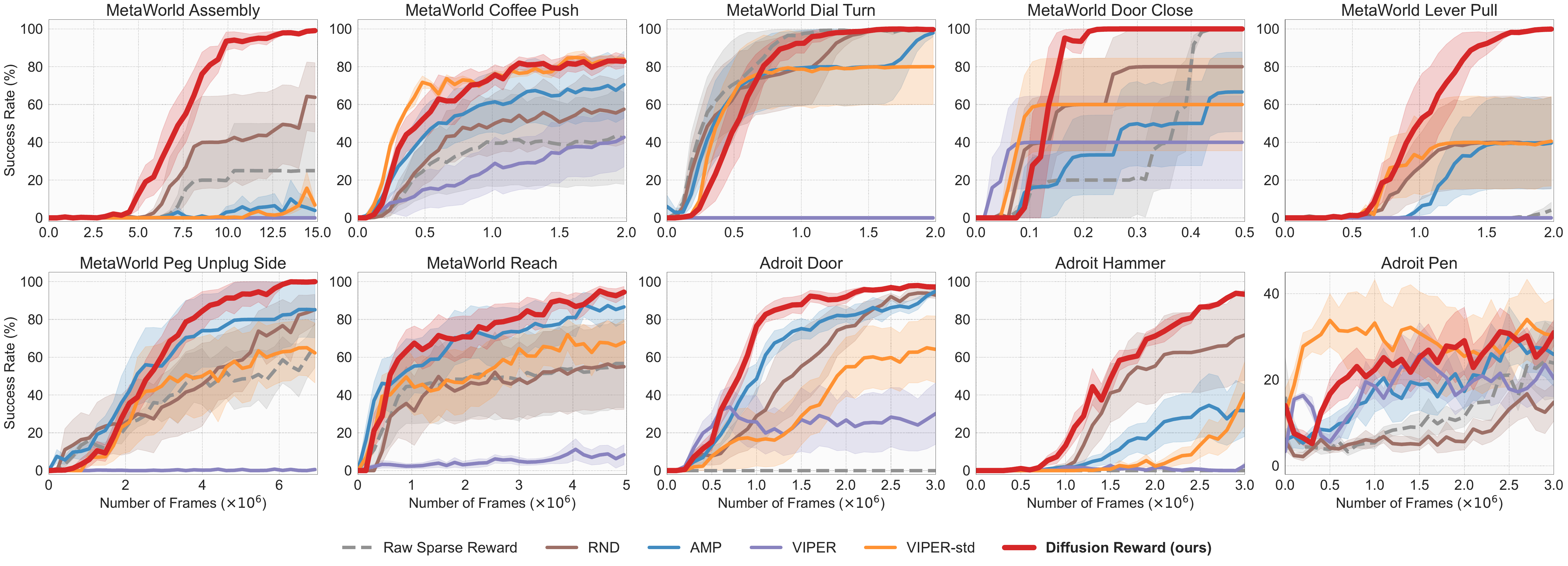}
\caption{\textbf{Main results.} Success rate for our method and baselines on 7 gripper manipulation tasks from MetaWorld and 3 dexterous manipulation tasks from Adroit with image observations. Our method achieves prominent performance on all tasks, and significantly outperforms baselines on complex door and hammer tasks. Results are means of 5 runs with the standard error (shaded area).}
\vspace{-0.1in}
\label{fig:main}
\end{figure}

\subsection{Main Results}\label{subsec:main}

\paragraph{Comparison with non-pretraining methods.} We present the learning curves of success rates for each method over two simulation domains in Figure~\ref{fig:overview} and~\ref{fig:main}. The results show that solely using sparse task rewards enables progress in relatively straightforward tasks such as Reach and Dial Turn. However, it encounters significant challenges in more complex ones, exemplified by the Door and Hammer within the dexterous hand manipulation domain. The incorporation of pure novelty-seeking rewards (RND) unsurprisingly enhances the RL agent's exploration, particularly evident in addressing moderately complex tasks like Coffee Push and Assembly. Nevertheless, the performance on dexterous hand manipulation tasks is still unsatisfactory due to the lack of expert-instructed exploration at prohibitively large configuration space. Conversely, AMP explicitly employs expert-guided rewards to incentivize the exploration of expert-like behaviors. While it generally outperforms RND in simpler tasks, its efficacy diminishes in more complex ones such as Lever Pull and Hammer. 

\pskip
\paragraph{Comparison with reward pretraining method.} The above observations suggest that the combination of expert-instructed rewards and novelty-seeking rewards is likely to perform favorably. Despite the incorporation of such a combination in the VIPER reward, its empirical performance unexpectedly falls short of both RND and AMP. We posit this is attributed to the significantly varied scales of expert-instructed rewards (i.e., log-likelihood), which may eliminate the effect of novelty-seeking rewards. VIPER-std alleviates this issue and shows better performance than the non-standardized version, especially on Coffee Push and Pen, while still underperforming our method. These results are in line with two limitations outlined in Section~\ref{subsec:rew} and further verified in Figure~\ref{fig:e_ll}, indicating VIPER's struggle in capturing complex video distributions within intricate tasks. In sharp contrast to VIPER, our proposed method not only utilizes the modeling capabilities of diffusion models, but also uses conditional entropy as a reward function to accelerate exploration. The results showcase remarkable performance improvements of \textbf{38\%} and \textbf{35\%} over the best-performing baselines with the same training steps across MetaWorld and Adroit, respectively.


\begin{figure}[!t]
  \centering
  \includegraphics[width=1\linewidth]{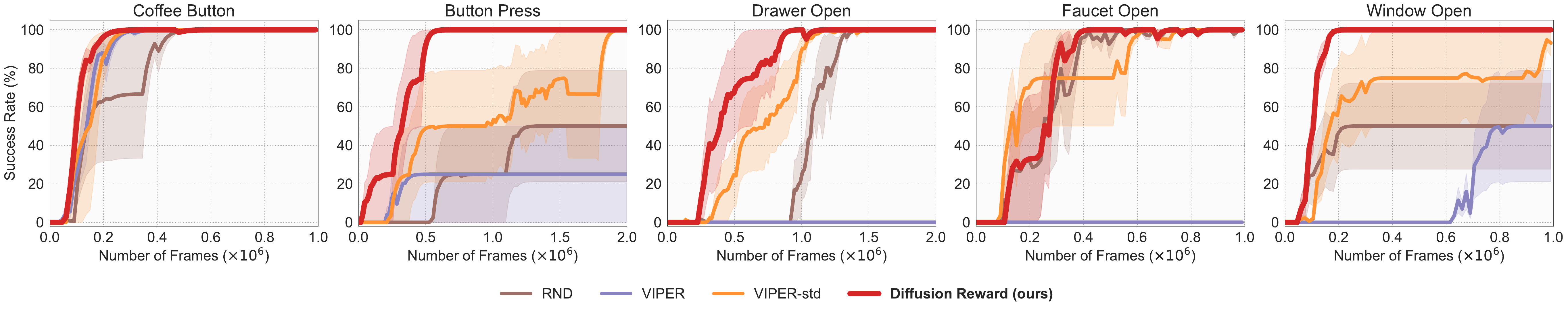}
\caption{\textbf{Success rate curves on 5 unseen MetaWorld tasks.} Diffusion Reward could generalize to unseen tasks directly and produce reasonable rewards, largely surpassing other baselines. Results are means of 4 runs with the standard errors (shaded area).}
\vspace{-0.05in}
\label{fig:generalization}
\end{figure}

\subsection{Zero-shot Reward Generalization}\label{subsec:generalization}
\paragraph{Reward analysis.} The advances of video diffusion models in generating samples beyond their training data, as exemplified in text-to-image video generation~\cite{ho2022imagen, esser2023structure}, motivates our exploration into the potential of \ours to generalize to previously unseen tasks. To investigate this, we first visuale the learned returns of trajectories with varying qualities derived from 15 diverse unseen tasks from MetaWorld (see Figure~\ref{fig:cond_entropy}). Expectedly, the distinctions between trajectories of varied qualities are less noticeable than those observed in tasks seen during pretraining. Nonetheless, our pretrained reward model still exhibits a consistent trend where expert-like behaviors receive relatively high learned returns, owing to the generalization prowess of the video diffusion model.

\pskip
\paragraph{RL performance.} Subsequently, we directly apply our pretrained rewards to 5 tasks (more in the Appendix) involving diverse objects without additional tuning. The outcomes, as shown in Figure~\ref{fig:generalization}, affirm that our reward effectively guides RL exploration and largely outperforms other baseline methods across all tasks. Notably, our approach proves helpful in constraining the exploration space of RND due to its retained ability to discriminate between expert-like and expert-unlike behaviors. Meanwhile, VIPER 
generalizes worse than ours on most tasks, especially on complex ones like opening objects, partially attributed to the combined limitations of the adopted video models and log-likelihood rewards. Here we exclude comparison with AMP because it learns reward online, which requires unavailable expert data on the target task. These results not only verify the efficacy of our method, but also point towards the potential of employing larger diffusion models and integrating other modalities (e.g., text-based task specification) to enhance the generalization capabilities of our approach further.

\begin{figure*}[t]

\begin{minipage}{0.46\linewidth}
    \centering
    \includegraphics[width=0.85\linewidth]{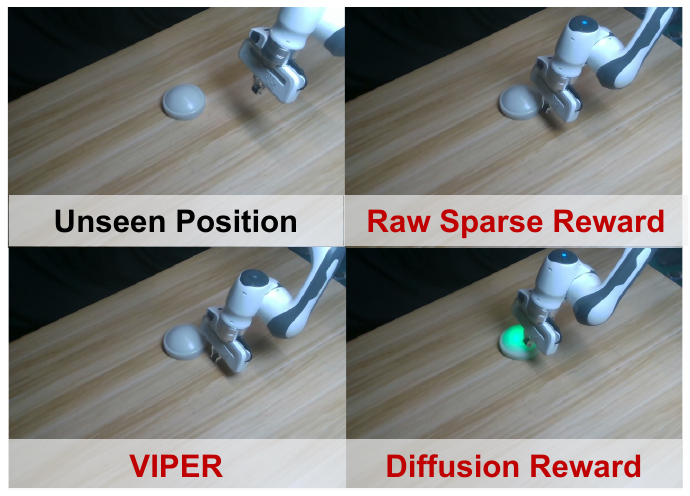}
    \caption{\textbf{Generalization in real-world task.} Policy trained with our method can successfully press the button at an unseen position.}
    \label{fig:real_generalization}
\end{minipage}
\hfill
\begin{minipage}{0.5\linewidth}
    \centering
    \includegraphics[width=0.85\linewidth]{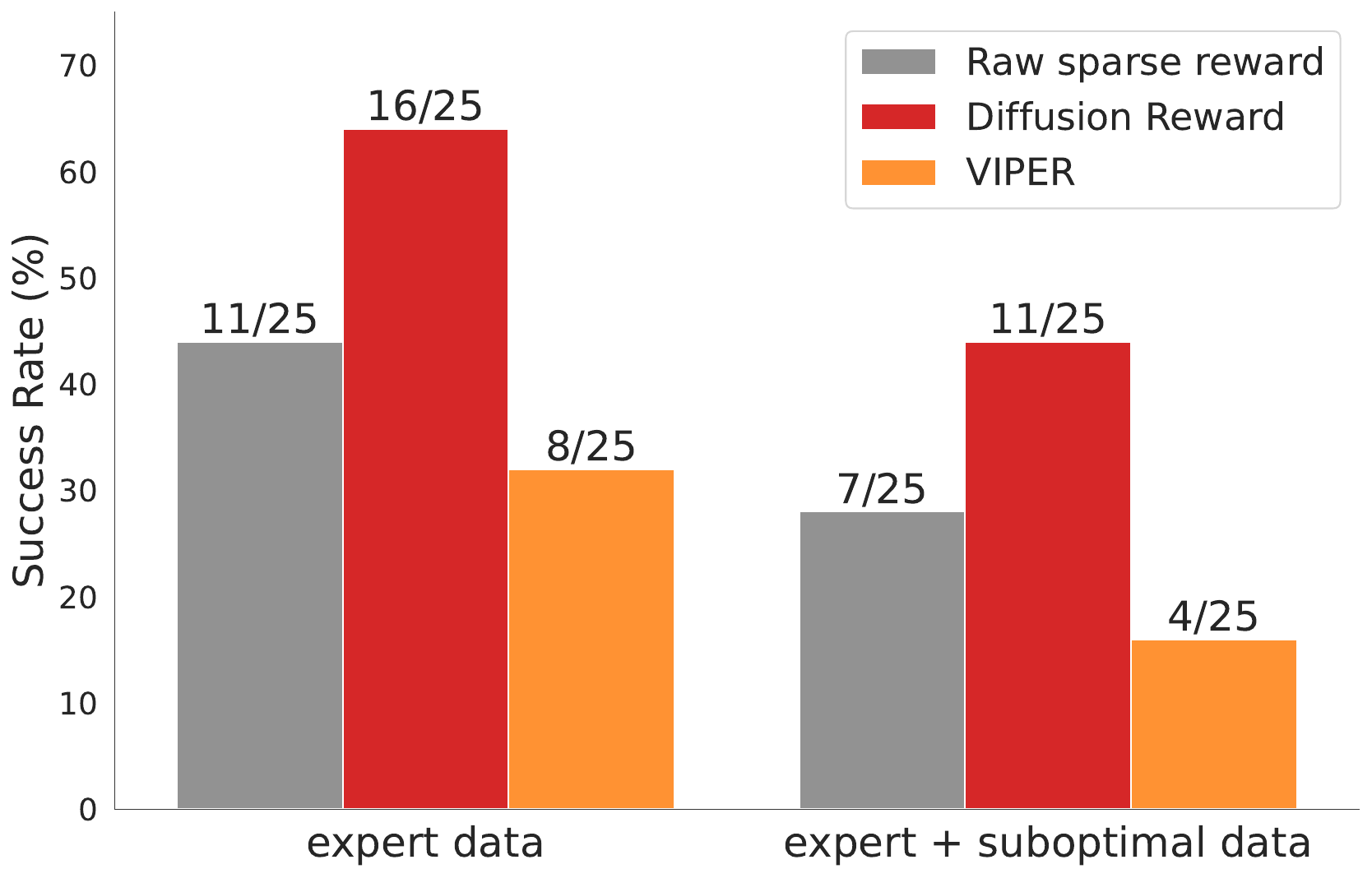}
    \caption{\textbf{Success rate of real-world offline RL.} Our method significantly outperforms baselines on the Button Press task given different offline data.}
    \label{fig:real_offline_rl}
\end{minipage}
\end{figure*}

\subsection{Real Robot Evaluation}\label{subsec:real-robot-videos}
\paragraph{Reward analysis.} We consider the real robot task that aims to pick up a bowl on the table. To train and test our reward model, we collect 20 real robot videos (10 expert and 10 random) with an Allegro hand, a Franka arm, and a RealSense, operated by humans. Visualization results in Figure~\ref{fig:reward-curve-real-robot} show that \ours can appropriately assign expert videos relatively higher rewards and, in contrast, random videos lower rewards, indicating the potential of our method for real-world robot manipulation tasks. We present more curves of multiple trials and reward analysis related to baselines in the Appendix. 

\paragraph{Offline RL performance.} We conduct offline RL evaluations in the Button Press task using a parallel gripper as the end-effector. Expert videos are used to train each reward model for subsequent offline RL training (see the Appendix for more details). Our results show that Diffusion Reward produces a more generalizable policy for pressing the button at an unseen position, as illustrated in Figure~\ref{fig:real_generalization}. Moreover, Figure~\ref{fig:real_offline_rl} quantitatively demonstrates that Diffusion Reward significantly outperforms VIPER in terms of real-world applicability.

\subsection{Ablation Studies}\label{subsec:abl}
\begin{figure*}[t]
  \centering
    \includegraphics[width=1\linewidth]{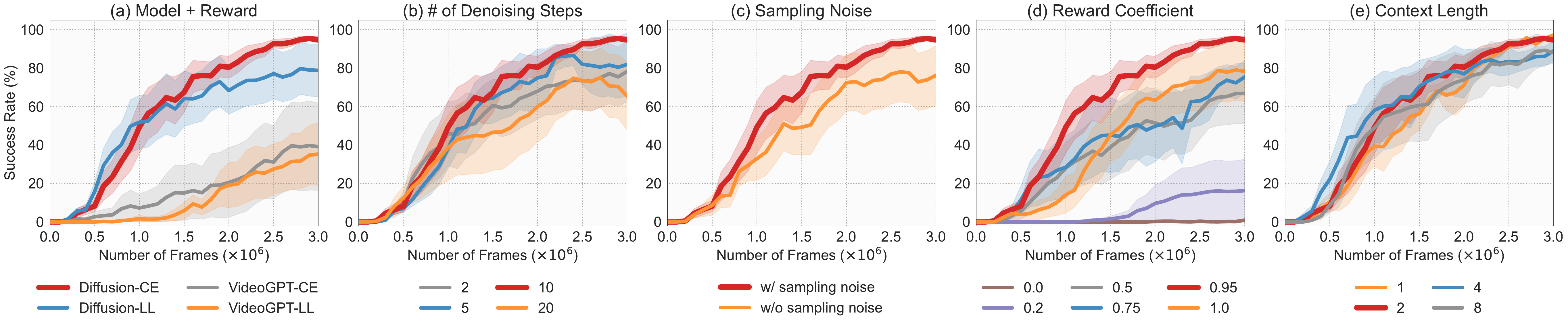}
    \caption{\textbf{Ablations.} Success rate curves for ablated versions of \ours, aggregated over Door and Hammer from Adroit. (a) We test the combinations of generative models and rewards to show the benefit of estimating conditional entropy with the diffusion model. (b) We ablate the choice of the number of denoising steps. (c) We demonstrate that the inherent randomness of \ours from the reverse process helps RL exploration. (d) We ablate the choice of reward coefficient. (e) We test the effect of the number of conditional frames. Results are means of 3 seeds with the standard error (shaded area). {\color{mred} \textbf{Red}} is our default.}
    \label{fig:abl_all}
\end{figure*}

As shown in Figure~\ref{fig:abl_all}, we ablate key design choices in our proposed framework in the previous experiments, aiming to reveal more insights into the quantitative performance of our method.  We present more detailed analyses below.

\pskip
\paragraph{Conditional entropy with diffusion model.}
As our method has demonstrated better performance than VIPER, 
we take a further step to see the joint effect of the reward types and the video prediction models. Specifically, we systematically evaluate all possible combinations of conditional entropy and log-likelihood as reward signals, each paired with either diffusion-based or transformer-based video prediction models. 
Note that the same vector-quantized encoder is used for all models, ensuring the variations are solely attributable to the chosen video prediction model and reward. 
The outcomes consistently align with the observations delineated in Figure~\ref{fig:e_ll}, indicating two-fold conclusions: (1) video diffusion models are more adept at capturing the complex distribution of expert videos in complex tasks, thus resulting in more informative rewards; (2) employing conditional entropy as rewards prove more productive in RL exploration than using log-likelihood, partially owing to the more generalizable reward inference on trajectories unseen in reward pretraining stage. 

\pskip
\paragraph{Denoising steps.} The number of timesteps involved in the reverse process governs the quality and diversity of generated frames~\cite{song2020denoising}. This study seeks to investigate its impact on the derived reward and subsequent RL performance by gradually increasing the number of denoising steps from 2 to 20. The findings indicate that an intermediate choice, approximately around 10 steps, achieves the best performance. This suggests that an intermediate choice balances generative quality and diversity well, thereby producing effective rewards for RL exploration. Furthermore, we observe that the speed of reward inference declines with an increase in denoising steps, suggesting the benefit of adopting advanced techniques to expedite the diffusion process if requiring more denoising steps. 

\pskip
\paragraph{Sampling noise in diffusion process.} We hypothesize that the randomness in the diffusion process can accelerate RL exploration, akin to the stochastic characteristics of maximum entropy RL~\cite{ziebart2008maximum}. To substantiate this point, we design a variant of Diffusion Reward wherein the sampling noise is deliberately set as 0 during the reverse process, ensuring the reward becomes deterministic given the identical historical observations. The outcomes show a discernible degradation in performance when employing a deterministic reward. Notably, this decline in performance aligns with results observed when combining the diffusion model with log-likelihood rewards, wherein the rewards are also deterministic. Consequently, our findings demonstrate that the inherent randomness of \ours from the reverse process indeed contributes to RL exploration.

\pskip
\paragraph{Reward coefficient $\alpha$.} The reward coefficient $\alpha$ determines the relative importance of the conditional entropy reward against novelty-seeking. We investigate the effect of this parameter by gradually decreasing the value of $\alpha$ from 1 to 0. The results show that $\alpha$ around 0.95 achieves the best performance, while too-large ones (akin to RND only) and too-small ones (akin to no RND) exhibit significant performance drops. This suggests the domination of Diffusion Reward may still result in getting stuck to local optima, while our proposed reward effectively helps the RL agent to narrow down the wide 
exploration space of novelty-seeking rewards. 

\pskip
\paragraph{Context length.} The number of historical frames determines the extent of temporal information being encoded during video diffusion, thus influencing the generating process of video diffusion and induced reward inference. To investigate its effect on downstream RL, we test different choices of context length. The results suggest that opting for 1 or 2 historical frames proves sufficient to generate highly effective rewards, owing to the robust generative capabilities inherent in the diffusion model. Interestingly, a marginal decline in performance is observed when the context length is extended to 4 or 8 frames. This phenomenon may be attributed to potential overfitting to expert trajectories, resulting in inferred rewards that exhibit suboptimal generalization to previously unseen trajectories.




%% file: sec/5_conclusion.tex
\section{Conclusion}

In this work, we propose \ours, a novel framework that extracts dense rewards from a pretrained conditional video diffusion model for RL tasks. We first pretrain a video diffusion model using expert videos and observe that the entropy of the predicted distribution well discriminates the expert-level trajectories and under-expert-level ones. Consequently, we use its standardized entropy, plus the exploration reward and the sparse environmental reward, as an informative reward signal. We evaluate \ours across 10 visual robotic manipulation tasks from MetaWorld and Adroit and observe prominent performance improvements over 2 domains. Additionally, we present results of reasonable rewards generated in real robot tasks and associated promising offline RL performance, highlighting the real-world applicability of our method. Furthermore, we demonstrate that our pretrained rewards can instruct RL to solve unseen tasks successfully and effectively, largely surpassing baseline methods. This underscores the potential of large-scale pretrained diffusion models in reward generalization. 

\paragraph{Limitation and future work.} Future work will leverage larger diffusion models sourced from a wider dataset to solve diverse simulation and real-world tasks. The incorporation of additional modalities, such as language, will be explored to augment the generalization capabilities of \ours. In addition, enhancing the diffusion-based reward itself, including strategies to balance entropy reward and exploration reward, the estimations of conditional entropy, and accelerating reward inference speed, holds promise for yielding better outcomes. 

%% file: sec/6_appendix.tex
\clearpage
\appendix
\section{Implementation Details} 
In this section, we provide further implementation details on \ours and baselines. Note that all methods use the same RL backbone and sparse task reward. We maintain all settings except reward pretraining (if exist) identical.

\subsection{\ours Implemenatation}
\paragraph{Codebase.} Our codebase of VQ-GAN~\cite{esser2021taming} is built upon the implementation in \url{https://github.com/dome272/VQGAN-pytorch}, which provides clean code structure and shows fast inference speed. The codebase of VQ-Diffusion~\cite{VQDiffusion} is built upon the official implementation, which is publicly available on \url{https://github.com/microsoft/VQ-Diffusion}. For the downstream RL, we adopt the official implementation of DrQv2~\cite{yarats2021mastering} as RL backbone, which is publicly available on \url{https://github.com/facebookresearch/drqv2}, and the implementation of RND as exploration reward available on \url{https://github.com/jcwleo/random-network-distillation-pytorch}. 

\pskip
\paragraph{Network architectures.} The major network architectures employed in \ours follow the original implementation provided by the codebase above (refer to corresponding papers for more details), except for modifications performed in the conditional video diffusion part. Specifically, each historical frame, encoded by the encoder $E$ and quantizer $Q$ learned with VQ-GAN, is tokenized ($8\times 8$) by the condition network, concatenated with others ($2\times 8\times 8$), fed into embedding networks with a dimension of 1024. The resulting condition embedding with a dimension of $128\times 1024$ is passed to subsequent conditional diffusion. 

\pskip
\paragraph{Hyperparameters.} We list the important hyperparameters of VQ-GAN, VQ-Diffusion, DrQv2 with \ours in Table~\ref{table:vqgan},~\ref{table:vqdiffusion}, and~\ref{table:drqv2}, respectively. 

\pskip
\paragraph{Entropy estimation details.} As described in Section~\ref{subsec:rew}, the variational bound of conditional entropy in Eq.~\eqref{eq:ce} can be estimate by Eq.~\eqref{eq:estimation}. Such estimation is realized with the closed-form distribution~\cite{sohl2015deep,kingma2013auto} (e.g., discrete multivariate distribution) in this work. Specifically, the variational bound of conditional log-likelihood in Eq.~\eqref{eq:ce} can be simplified following~\cite{sohl2015deep}, resulting in our estimation of entropy reward $r^\mathrm{ce}$ as follows:
\begin{equation}
\begin{split}
    &r^{\mathrm{ce}}(\bm{x}_{k-1}) = \frac{1}{M}\sum_{j=1}^M \big( \log  p_\theta(\tilde{\bm{z}}_{k}^0|\tilde{\bm{z}}_{k}^1,\bm{z}_{c}) \\ 
    &+ \sum_{t=1}^{T-1} D_{KL}(q(\tilde{\bm{z}}_{k}^{t-1}|\tilde{\bm{z}}_{k}^{t}, \tilde{\bm{z}}_{k}^{0})\| p_\theta(\tilde{\bm{z}}_{k}^{t-1}|\tilde{\bm{z}}_{k}^{t},\bm{z}_{c})) \\
    &+ D_{KL}(q(\tilde{\bm{z}}_{k}^{T}|\tilde{\bm{z}}_{k}^{0})\|p(\tilde{\bm{z}}_k^T) \big),
\end{split}
\end{equation}
where $D_{KL}$ denotes the Kullback–Leibler divergence, $p(\tilde{\bm{z}}_k^T)$ follows the prior distribution of random noise at timestep $T$, and $\tilde{\bm{z}}_{k}^{0:T}\sim p_\theta(\bm{z}_k^{0:T}|\bm{z}_c)$ represents the denoised samples via inverse diffusion process. We set $M=1$ to ensure high reward inference speed while retaining its discrimination on expert-like and -unlike behaviors. We present more analyses in Section~\ref{sec:more_abl}.

\begin{figure}[t]
  \centering
    \includegraphics[width=1\linewidth]{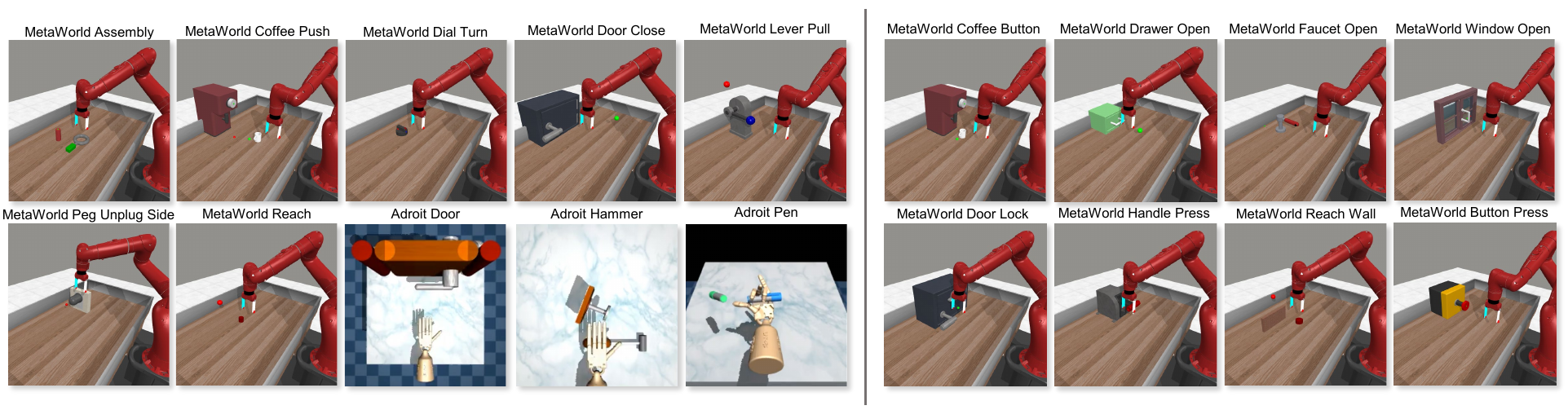}
    \vspace{-0.275in}
    \caption{\textbf{Task descriptions.} (\textit{left}) 10 seen training tasks from MetaWorld and Adroit. (\textit{right}) 8 unseen tasks from MetaWorld.}
    \label{fig:task_description}
    \vskip -0.1in
\end{figure}

\subsection{Baselines Implementations}
\paragraph{RND implementation.} This baseline combines the sparse environmental reward with the RND exploration reward, which is equivalent to setting the reward coefficient $\alpha$ as 1 in \ours. To this end, we implement this baseline by simply removing the entropy reward and keeping other settings identical.  

\pskip
\paragraph{VIPER implementation.} We implement their adopted VideoGPT based on the official code provided in  \url{https://github.com/wilson1yan/VideoGPT} with clean GPT implementation from \url{https://github.com/karpathy/minGPT}. The calculation of video prediction (log-likelihood) rewards follows the official JAX implementation provided in \url{https://github.com/Alescontrela/viper_rl}, which uses 'teacher-forcing' practice (where ground truth context is provided for each step) for fast inference speed. 

\pskip
\paragraph{AMP implementation.} The implementation is based on the official code in \href{https://github.com/xbpeng/DeepMimic}{https://github.com/xbpeng/DeepMimic}. 
The encoder consists of three 32-channel convolutional layers interpolated with \texttt{ReLU} activation. The discriminator is implemented as a 3-layer MLP with hidden dimensions of 256 and \texttt{Tanh} activation. 

\subsection{Real-world Offline RL Implementations}
We evaluate vision-based offline RL in a real-world Button Press task using a Franka robotic arm with a parallel gripper. We gather 50 expert demonstration videos (rescale to $84\times 84$ resolution) to train reward models for relabeling the offline dataset, which is then used to train policies with the vision-based offline RL method, DrQ+BC~\cite{luchallenges}. We follow their default hyperparameters. The suboptimal data contains 50\% randomly-taken actions.

\section{Task Descriptions}
We select 7 gripper manipulation tasks from MetaWorld~\cite{Yu2019MetaWorldAB} and 3 dexterous manipulation tasks from Adroit~\cite{Rajeswaran2017LearningCD}, as visualized in Figure~\ref{fig:task_description}. The tasks are widely used in visual RL and are chosen to be diverse in objects and manipulating skills. All tasks render $64\times 64$-dimensional RGB as the agent's observation and produce sparse environmental rewards. According to task complexity, we collect 20 expert videos for each MetaWorld task and 50 for each Adroit task. We describe each task below:
\begin{itemize}
    \item \texttt{Assembly} (MetaWorld, $\mathcal{A}\in \mathbb{R}^{4} $): the task is to pick up a nut and place it onto a peg with the gripper.
    \item \texttt{Coffee Push} (MetaWorld, $\mathcal{A}\in \mathbb{R}^{4} $): the task is to push a mug under the coffee machine to a target position with the gripper.
    \item \texttt{Dial Turn} (MetaWorld, $\mathcal{A}\in \mathbb{R}^{4} $): the task is to rotate the dial with the gripper.
    \item \texttt{Door Close} (MetaWorld, $\mathcal{A}\in \mathbb{R}^{4} $): the task is to close the door with the arm.
    \item \texttt{Lever Pull} (MetaWorld, $\mathcal{A}\in \mathbb{R}^{4} $): the task is to pull a lever up with the arm.
    \item \texttt{Peg Unplug Side} (MetaWorld, $\mathcal{A}\in \mathbb{R}^{4} $): the task is to unplug a peg sideways with the gripper.
    \item \texttt{Reach} (MetaWorld, $\mathcal{A}\in \mathbb{R}^{4} $): the task is to reach a target position with the end effector.
    \item \texttt{Door} (Adroit, $\mathcal{A}\in \mathbb{R}^{28} $): the task is to open the door to touch the door stopper.
    \item \texttt{Hammer} (Adroit, $\mathcal{A}\in \mathbb{R}^{26} $): the task is to pick up the hammer to hit the nail into the board. 
    \item \texttt{Pen} (Adroit, $\mathcal{A}\in \mathbb{R}^{18} $): the task is to reorient the pen in-hand to a target orientation. 
\end{itemize}

\begin{figure}[t]
  \centering
    \includegraphics[width=0.5\linewidth]{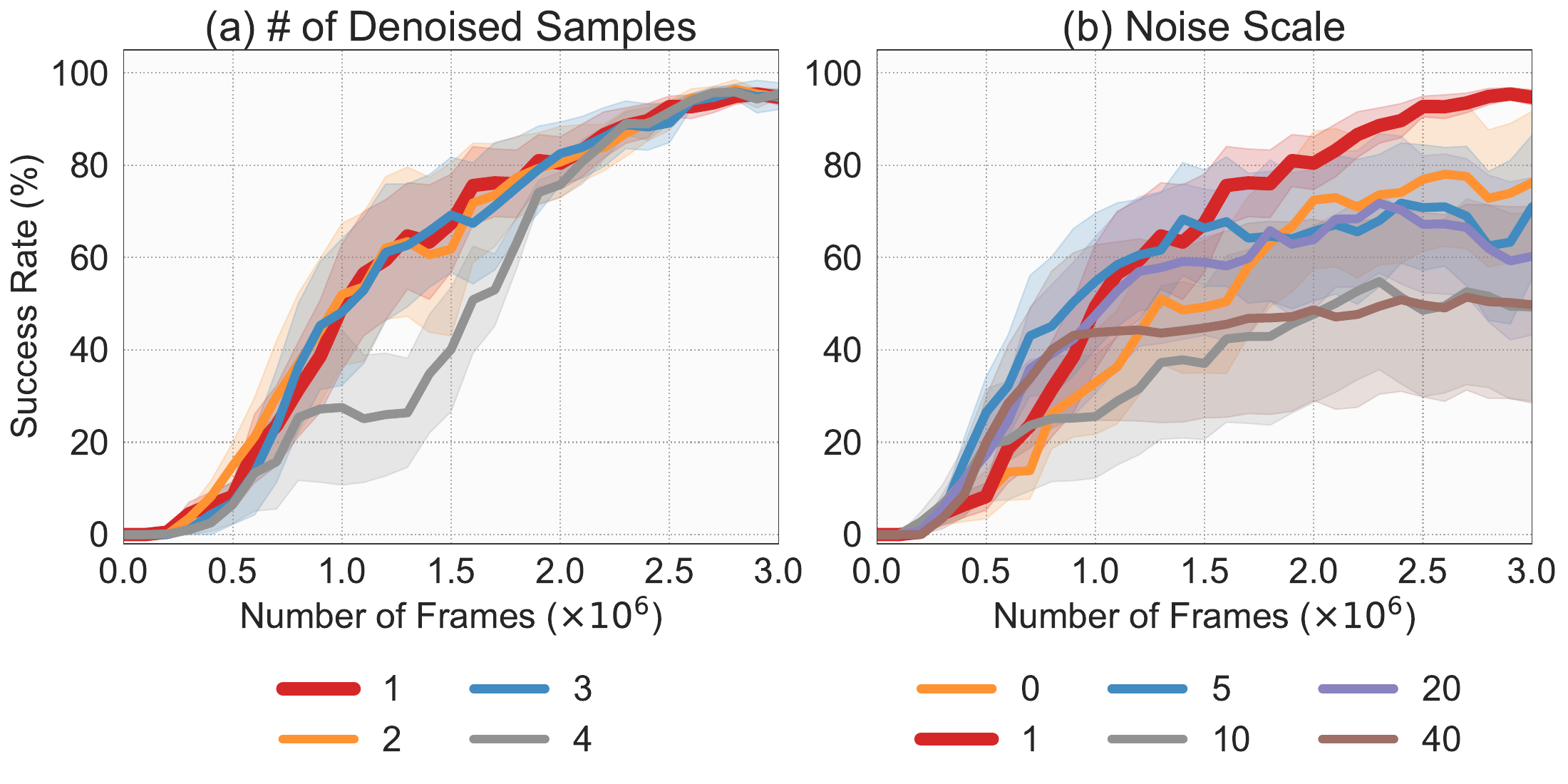}
    \caption{\textbf{Ablations on diffusion process.} The results in (a) suggest that overmuch denoised samples (i.e., 4) may hinder the exploration due to the low variance of estimated entropy. This is further verified in (b) where an appropriate choice of sampling noise scale results in more productive explorations. Results are means of 3 seeds with std. error (shaded area).{ \color{mred} \textbf{Red}} is our default.}
    \vskip -0.1in
    \label{fig:abl_denoise2}
\end{figure}

\section{More Video Prediction Results and Reward Analysis}
\subsection{Video Prediction Results}
\begin{figure*}[t]
  \centering
    \includegraphics[width=0.9\linewidth]{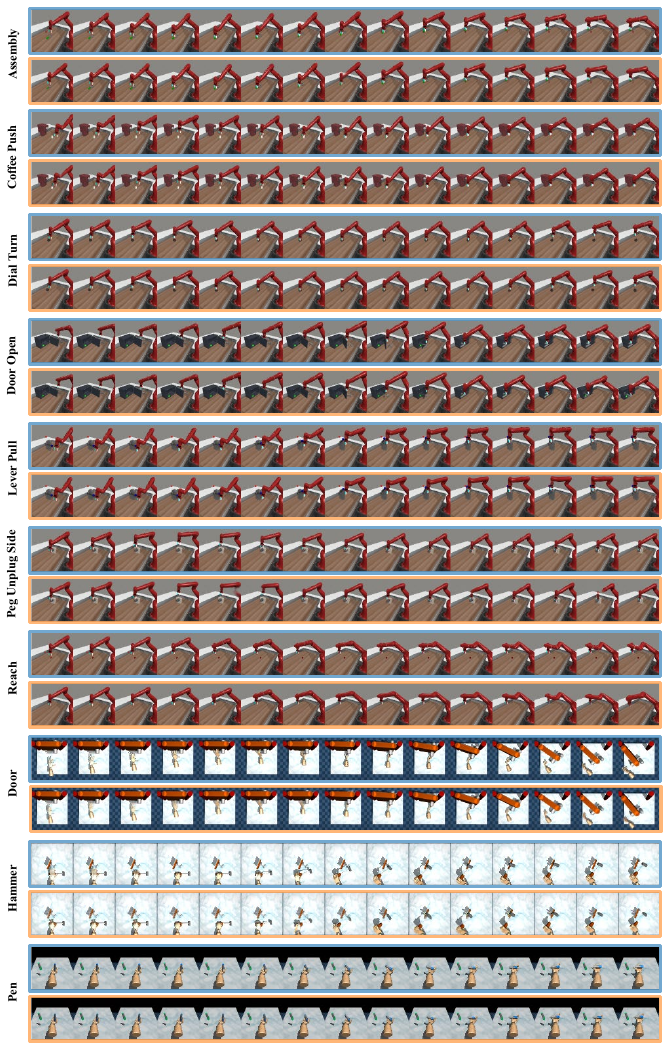}
    \caption{\textbf{Video prediction results.} Ground truth has {\color{mblue}blue} borders and prediction has {\color{morange}orange} borders.}
    \label{fig:all_video_prediction_results}
\end{figure*}

\paragraph{Qualitative results.}
We present the comparison between expert videos and prediction results in Figure~\ref{fig:all_video_prediction_results}. We find that the adopted video diffusion model can capture the complex distribution of expert videos from pretraining data and generalize well to unseen expert videos. Interestingly, we also observe that the colored target points in Door Close and Reach are sometimes mispredicted, which may explain the relatively slow exploration at the initial RL training stage, suggesting that a more powerful video diffusion model could be used to improve the reward quality further. 

\pskip
\paragraph{Quantitative analysis.} We compare the video prediction quality between \ours and VIPER in terms of three video metrics, SSIM, PSNR, and LPIPS. Results are shown in Table~\ref{tab:quant-video-prediction}, verifying that Diffusion models hold a stronger generalization ability on unseen real robot and simulation trajectories than VideoGPT and thus produce more informative rewards. 

\subsection{Reward Analysis}
\begin{wrapfigure}{r}{0.3\linewidth}
  \centering
    \vspace{-0.31in}
    \includegraphics[width=1\linewidth]{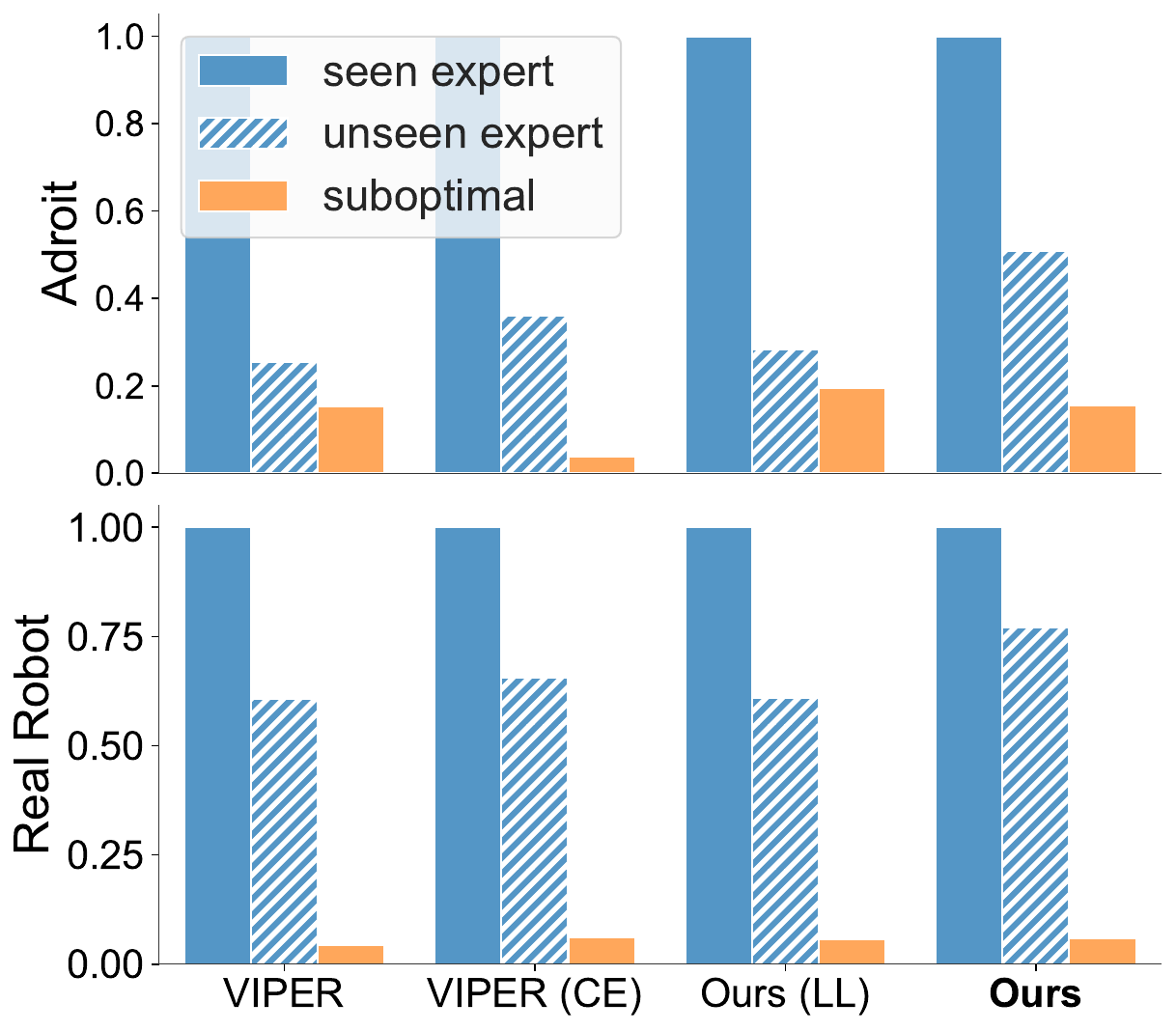}
    \vspace{-0.3in}
    \caption{\textbf{Reward analysis}.}
    \label{fig:e_ll_more}
    \vspace{-0.3in}
\end{wrapfigure}
In Figure~\ref{fig:e_ll}, we conclude that the conclusion that our rewards can generally discern unseen expert behaviors in the MetaWorld domain. We extensively strengthen this conclusion in the Adroit domain and real robot tasks with noisier human-operated demonstrations, as demonstrated in Figure~\ref{fig:e_ll_more}. Interestingly, we observe that the suboptimal behaviors are assigned with higher rewards by the diffusion model than by VideoGPT in the Adroit domain, which slightly differs from the observations in the MetaWorld domains. We posit that multiple reasons can be attributed to it. On the one hand, the suboptimal behaviors in the Adroit domain may differ from the ones in MetaWorld due to the higher action space. Besides, the scenes, including background and objects, vary dramatically in the Adorit tasks. Nonetheless, the entropy-based reward performs better than log-likelihood whichever video model is used.

\begin{figure*}[t]
  \centering
    \includegraphics[width=1.0\linewidth]{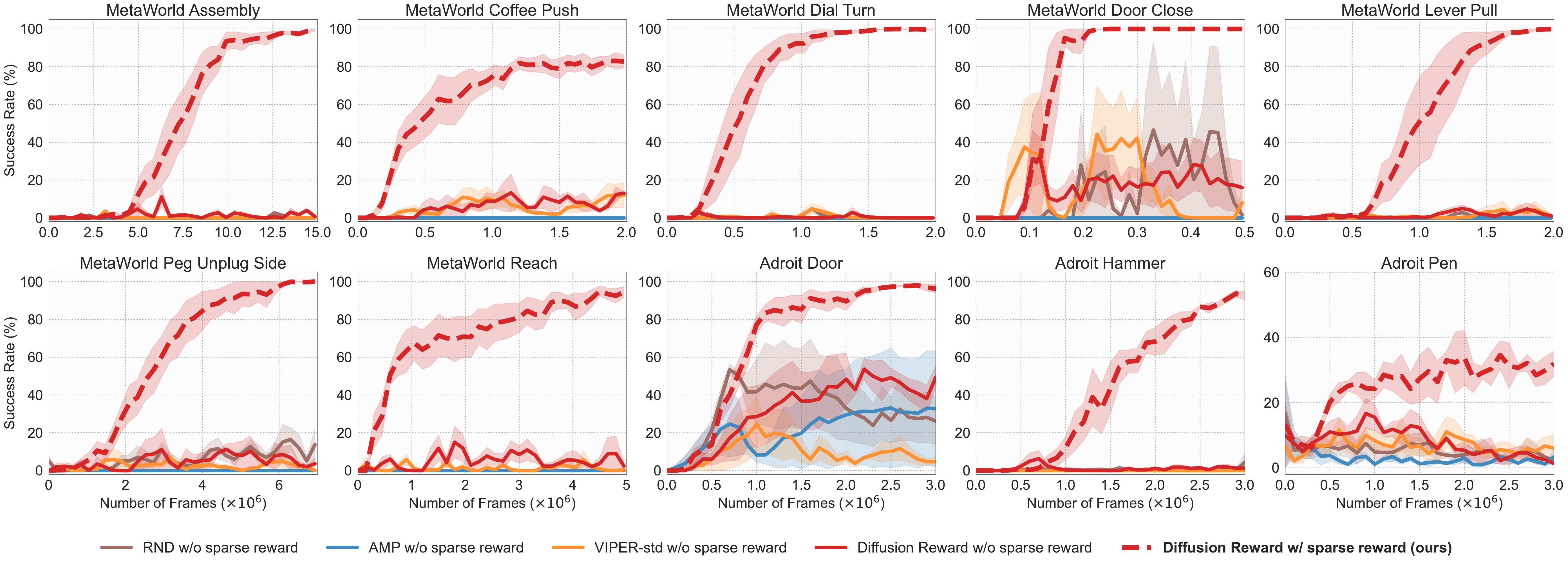}
    \vskip -0.1in
    \caption{\textbf{Effect of sparse environmental reward.} We demonstrate that incorporating sparse environmental reward as a task completion signal is necessary for solving complex manipulation tasks. Results are means of 3 seeds with standard errors (shaded area).}
    \label{fig:abl_sparse}
    \vskip -0.1in
\end{figure*}

\section{More Experiments}
\subsection{More Ablations}\label{sec:more_abl}
We conduct more ablations on our proposed \ours in this section. Results are aggregated over Door and Hammer (by default) from Adroit with 3 random seeds. 

\pskip
\paragraph{Sparse environmental reward.} The sparse environmental reward $r^\mathrm{spar}$ is integrated into Eq.~\eqref{eq:final_reward}, as the environmental supervision of completion is helpful for RL. We remove $r^\mathrm{spar}$ to study its effect. The results in Figure~\ref{fig:abl_sparse} show that the performances decrease dramatically without sparse environmental rewards, indicating that the signal of task completion is necessary for solving complex manipulation tasks. Interestingly, more favorable performance is observed in the Door task. We attribute this to the overlap supervision of RND reward and sparse reward, i.e., exploring novel states (door opening) is partially equivalent to completing the task. 

\pskip
\paragraph{Diffusion process.} Recall that, in Eq.~\eqref{eq:estimation}, we perform inverse process for $M$ times, and use the generated $M$ samples to estimate the conditional entropy as rewards. The results in Figure~\ref{fig:abl_denoise2}(a) show that the effect of the number of denoised samples has a slight influence on RL performance when increasing from 1 to 3. However, the learning progress gets stuck in the middle state with 4 denoised samples, though the asymptotic performance is still satisfactory. We posit that this is due to the low variance of estimated entropy, which may lead to more weight on exploitation instead of exploration.

To verify our hypothesis, we make a further ablation on the scale of sampling noise (e.g., uniform distribution) during the diffusion process. Different from Figure~\ref{fig:abl_all}(b), we gradually increase the sampling noise scale from 0 to 40. The results in Figure~\ref{fig:abl_denoise2}(b) indicate that an excessively low noise scale (e.g., 0) will bring low randomness of learned reward, resulting in more exploitative behaviors, and too high noise scale may produce more random explorations. In contrast, an intermediate choice of noise scale will bring an appropriate variance of estimated entropy, contributing to productive explorations. 

Meanwhile, we present the time efficiency of different numbers of diffusion processes with a single NVIDIA A40 in Table~\ref{table:time_efficiency}, where the Frames Per Second (FPS) decreases from 87.7 to 45.9 when the number of denoise samples increases from 1 to 4. It suggests that using 1 denoised sample is sufficient to provide informative rewards and retain high inference speed. 
\begin{table}[h]
\centering
\vspace{-0.17in}
\begin{tabular}{lcccc}
\toprule 
 & $M=1$ & $M=2$ & $M=3$ & $M=4$  \\
\midrule
FPS & 86.4 & 67.5 & 55.8 & 45.9 \\
\bottomrule
\end{tabular}
\vspace{0.08in}
\caption{Time efficiency of the number of samples $M$ on Door.}
\label{table:time_efficiency}
\vspace{-0.5in}
\end{table}

\begin{figure}[h]
    \vspace{-0.25in}
     \centering
     \begin{subfigure}[b]{0.25\linewidth}
         \centering
         \includegraphics[width=\textwidth]{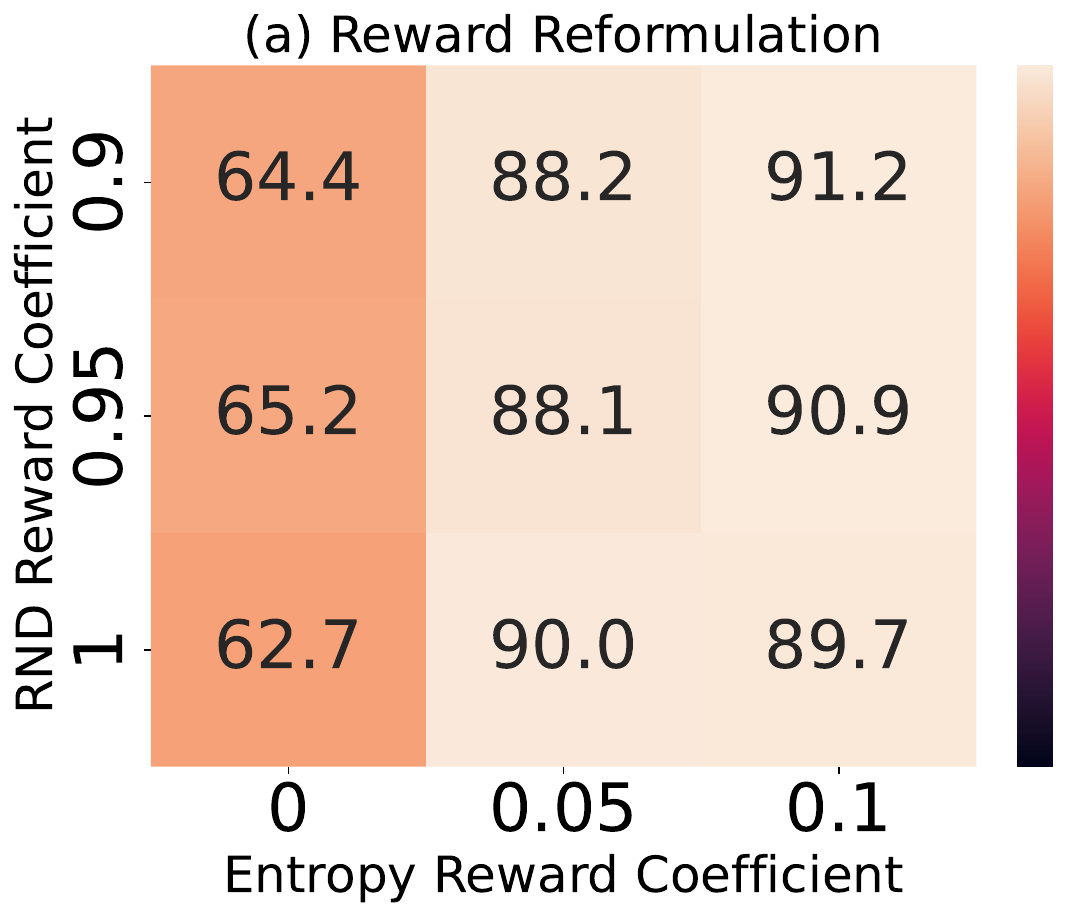}
     \end{subfigure}
     \begin{subfigure}[b]{0.25\linewidth}
         \centering
         \includegraphics[width=\textwidth]{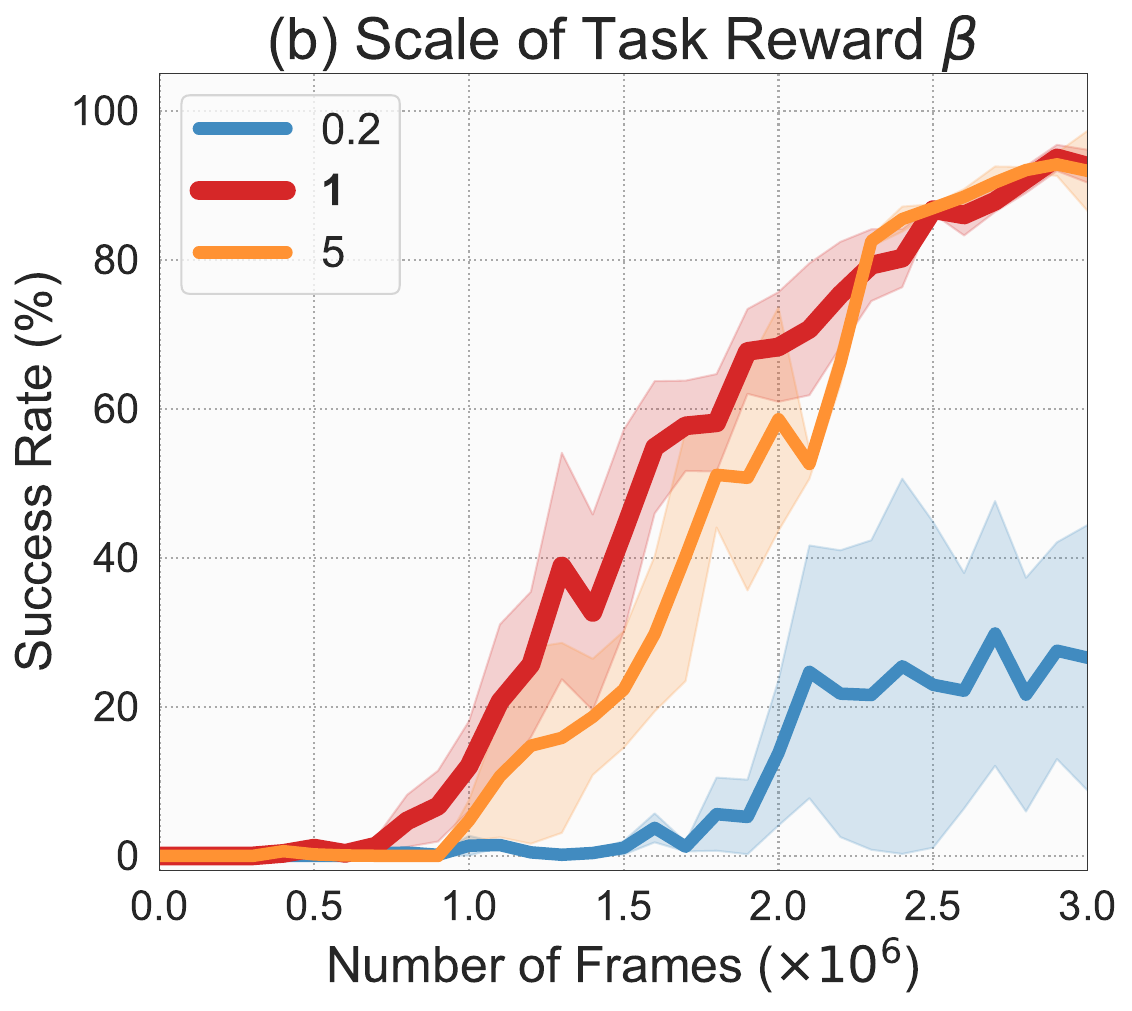}
     \end{subfigure}
     \vspace{-0.1in}
        \caption{\textbf{Reward coefficient.} The results in (a) suggest the introduction of an additional coefficient to balance entropy-based reward and RND reward has a negligible effect but brings more hyperparameter-tuning effort. The results in (b) suggest an appropriate scale of sparse task reward is crucial to give sufficient supervision signals.}
        \label{fig:reward_coef_analysis}
    \vspace{-0.25in}
\end{figure}

\pskip
\paragraph{Reward coefficient.} We introduce two coefficients to balance the effect of different reward components in Eq.~\eqref{eq:final_reward}. Firstly, we show that using a single coefficient $\alpha$ to balance $\bar{r}^{\mathrm{ce}}$ and $r^{\mathrm{rnd}}$ has no significant distinctions between using two independent ones in Figure~\ref{fig:reward_coef_analysis}(a). Secondly, we investigate the effect of the weight of sparse task reward $r^{\mathrm{spar}}$ in Figure~\ref{fig:reward_coef_analysis}(b). The results suggest that an appropriate coefficient of task reward $\beta$ is crucial, while a too-low choice may fail to provide sufficient task completion signals to the agent. We set $\beta=1$ throughout the whole paper which is appropriate enough.

\pskip
\begin{wrapfigure}{r}{0.275\linewidth}
  \centering
    \vspace{-0.31in}
    \includegraphics[width=1\linewidth]{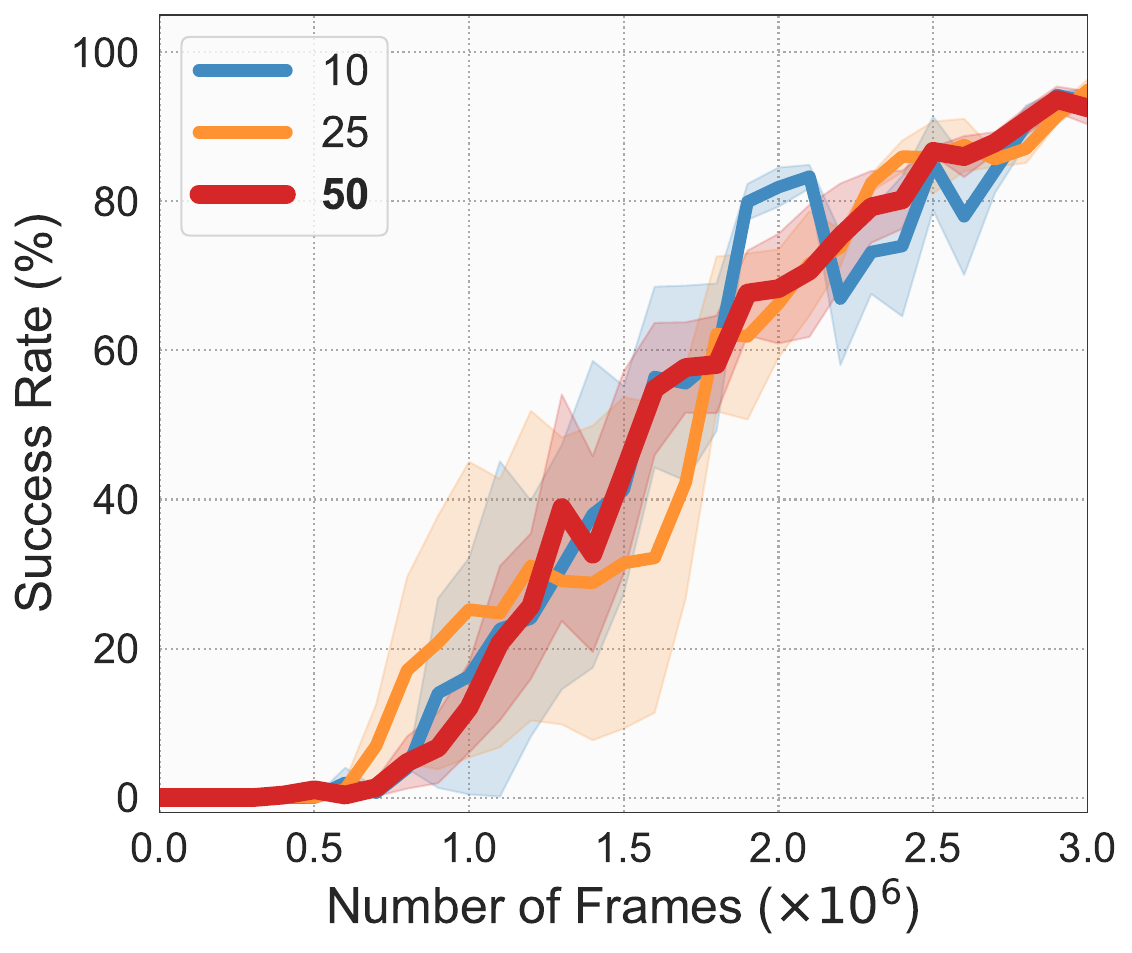}
    \vspace{-0.3in}
    \caption{\textbf{Demonstration amount} with 3 runs.}
    \label{fig:demo_num}
    \vspace{-0.3in}
\end{wrapfigure}
\paragraph{Number of demonstrations.} We investigate how sensible our method is to the number of expert videos in Figure~\ref{fig:demo_num}. Here we collect different numbers of expert videos from three selected Adroit tasks to train our reward models, and test the performance on the most complex Hammer task. Thanks to the strong modeling ability of diffusion models and the appropriate design of entropy-based rewards, the performance is still favorable even given 10 demonstrations for each task.

\subsection{More Comparison}
\pskip
\begin{wrapfigure}{r}{0.325\linewidth}
  \centering
  \vspace{-0.3in}
    \includegraphics[width=1\linewidth]{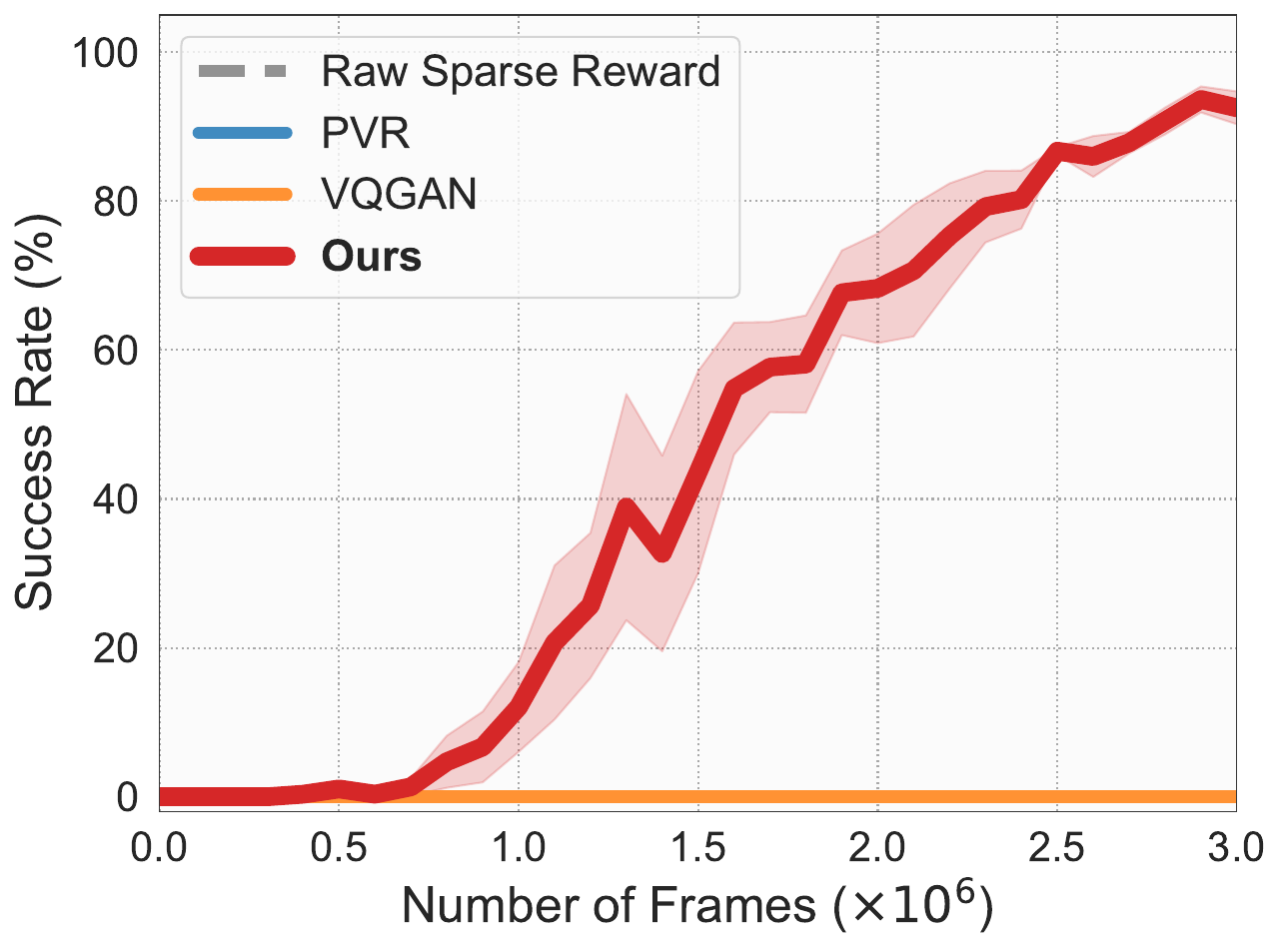}
    \vspace{-0.28in}
    \caption{\textbf{Comparison with visual representation pretraining} on Hammer task with 5 runs.}
    \label{fig:pretrain}
    \vspace{-0.27in}
\end{wrapfigure}
\paragraph{Comparison with other pretraining schemes.} To further demonstrate the benefit of pre-training our proposed rewards from videos, we do more comparisons with the scheme that pre-trains visual encoders from videos~\cite{xiao2022masked,nair2023r3m,parisi2022unsurprising}. Specifically, we evaluate two methods, including the representative one that uses PVR~\cite{parisi2022unsurprising} to pre-train RL encoder and the one that uses VQ-GAN instead. The results in Figure~\ref{fig:pretrain} suggest the remained difficulty of vision-based manipulation tasks even with pretrained visual encoders and, more importantly, the superiority of our reward-pretraining method. Meanwhile, there exist works that directly pre-train decent policies from data for RL~\cite{nair2020awac,lee2022offline,kun2023uni,pertsch2021accelerating, singh2020parrot, rao2021learning}. However, they are at the sacrifice of additional collections of reward labels, action labels, or task-agnostic data, while we learn rewards as priors from action-free expert-only videos.

\subsection{More Generalization Experiments}
\begin{figure}[t]
  \centering
    \includegraphics[width=0.75\linewidth]{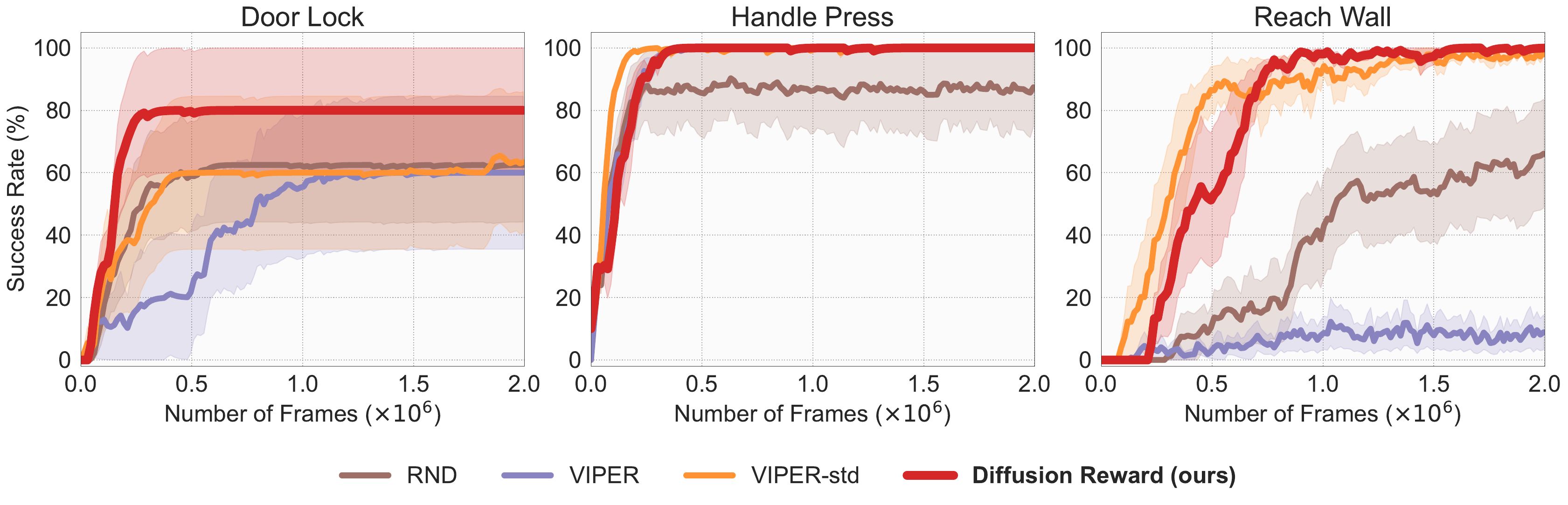}
    \vskip -0.1in
    \caption{\textbf{More results of reward generalization.} \ours exhibit better generalization ability than VIPER. Results are means of 5 seeds with 95\% standard errors (shaded area).}
    \label{fig:more_generalization}
    \vskip -0.1in
\end{figure}

\paragraph{More tasks.} Apart from 5 unseen tasks in Figure~\ref{fig:generalization}, we randomly select 3 more unseens tasks from MetaWorld to verify the zero-shot generalization capability of \ours. The result in Figure~\ref{fig:more_generalization} demonstrates that our method significantly outperforms VIPER on most tasks. Notably, our method also outperforms RND in Reach Wall in terms of productive exploration at the initial training stage. In the future, we will investigate the possibility of incorporating other modalities (e.g., text embedding for task description) to enhance the generalization ability of \ours. 

\pskip
\begin{wrapfigure}{r}{0.325\linewidth}
  \centering
    \vspace{-0.31in}
    \includegraphics[width=1\linewidth]{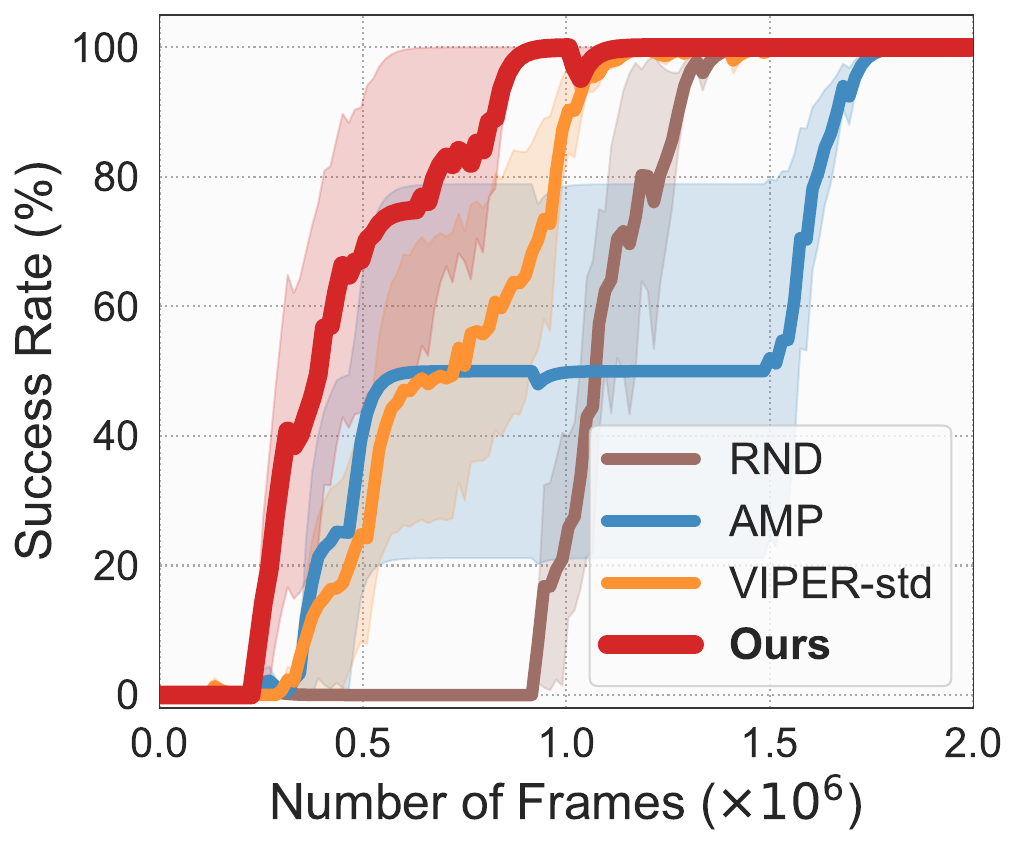}
    \vspace{-0.3in}
    \caption{\textbf{AMP generalization}}
    \label{fig:amp_gene}
    \vspace{-0.3in}
\end{wrapfigure}
\paragraph{Comparison with AMP.} Recall that AMP is excluded from our generalization experiments due to its requirement on target expert data. That said, we design a pseudo-generalization test in Figure~\ref{fig:amp_gene}, where AMP is trained on Door Close and frozen in the Drawer Open. We named it `pseudo' because, though AMP does not require expert videos from the target task (Drawer Open) here, it requires additional online interactions on Door Close tasks, which differs from our default settings.  We surprisingly observe a fair performance, possibly because AMP learns to discern the behavior of reaching objects and transfer this ability. Still, our method outperforms others.

\section{Visualization of Reward and Trajectory in Simulation and Real Robot}
\begin{figure*}[t]
     \centering
     \begin{subfigure}[b]{0.496\linewidth}
         \centering
         \includegraphics[width=1\linewidth]{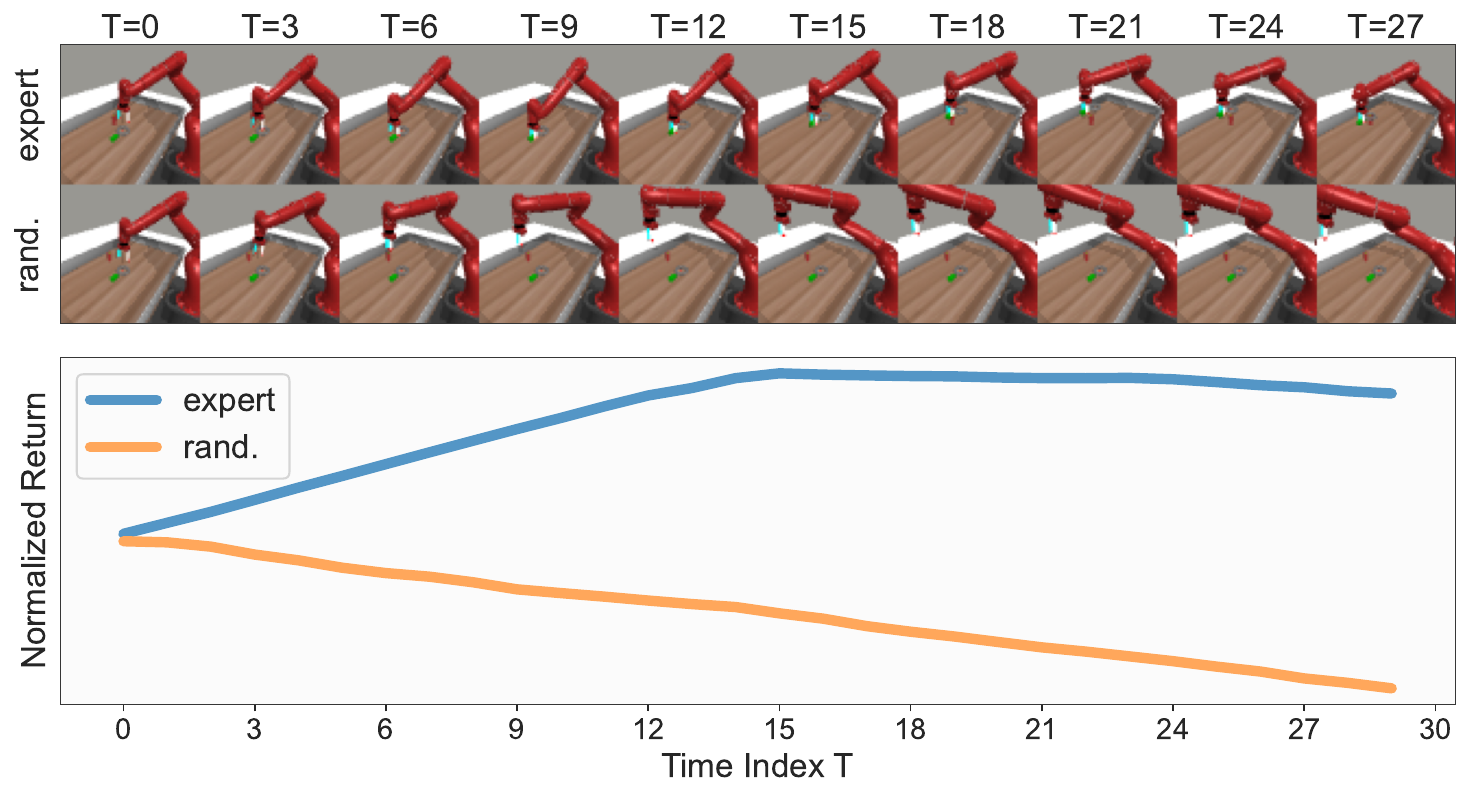}
     \end{subfigure}
     \begin{subfigure}[b]{0.496\linewidth}
         \centering
         \includegraphics[width=1.0\linewidth]{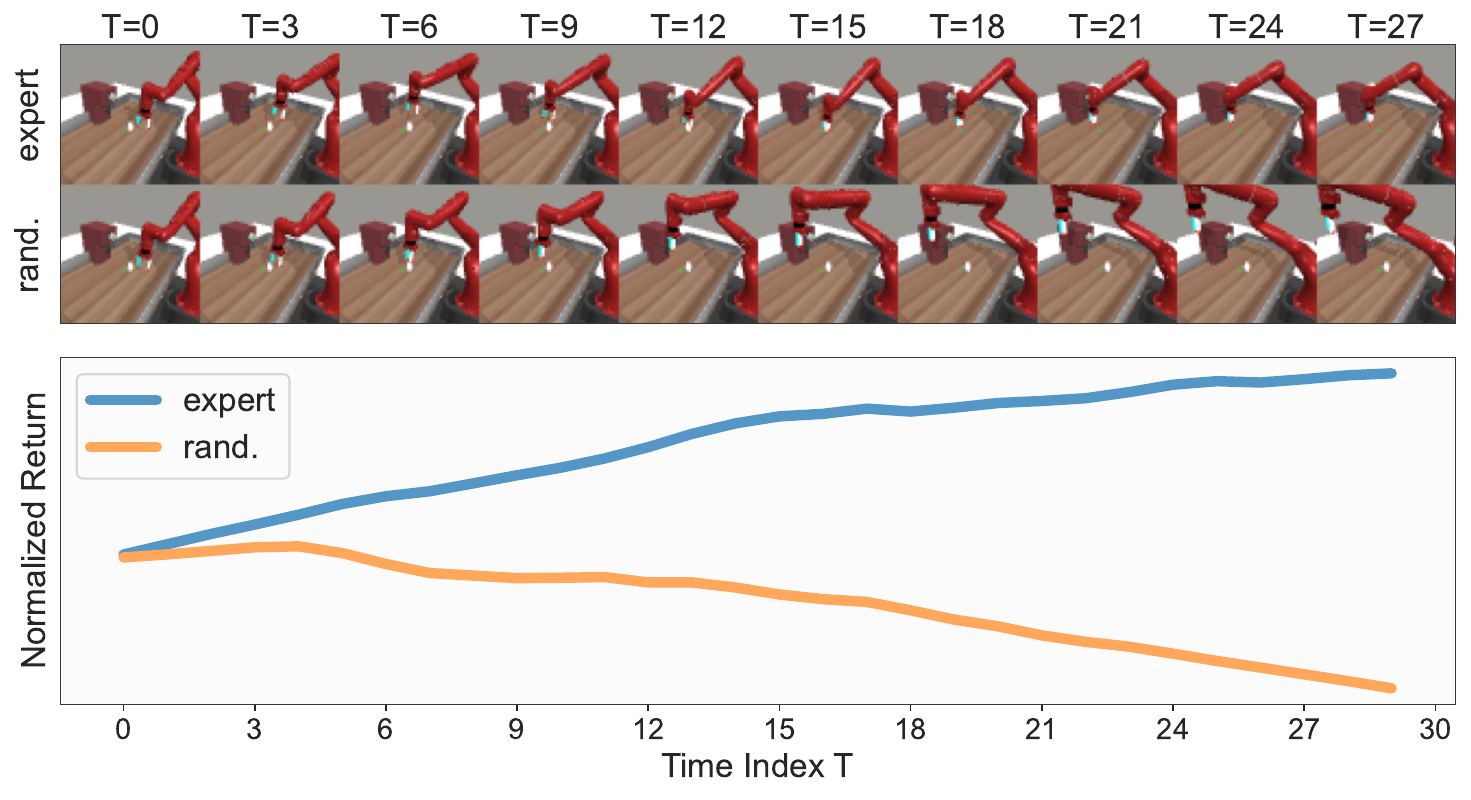}
     \end{subfigure}
     \\
     \begin{subfigure}[b]{0.496\linewidth}
         \centering
         \includegraphics[width=1.0\linewidth]{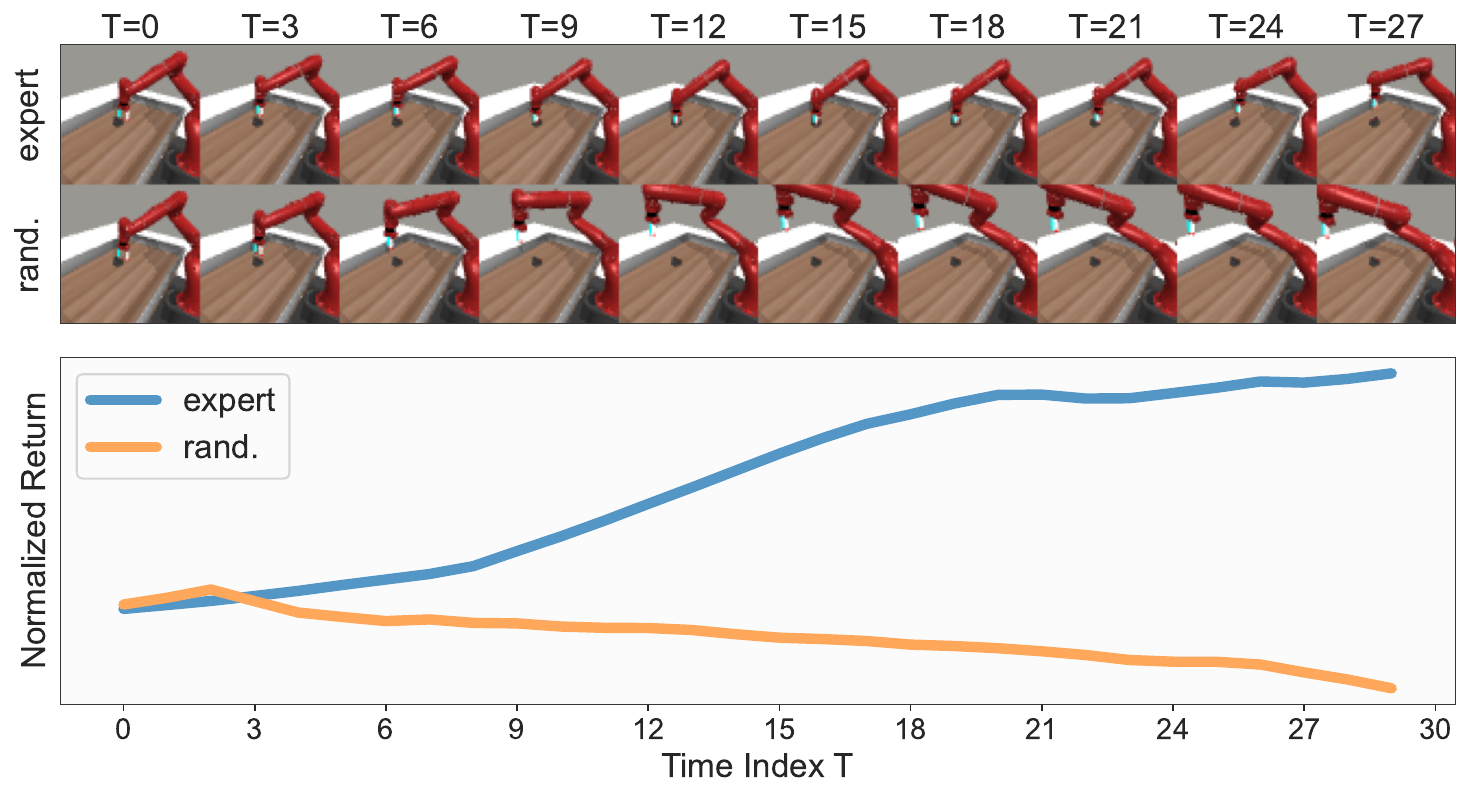}
     \end{subfigure}
     \begin{subfigure}[b]{0.496\linewidth}
         \centering
         \includegraphics[width=1.0\linewidth]{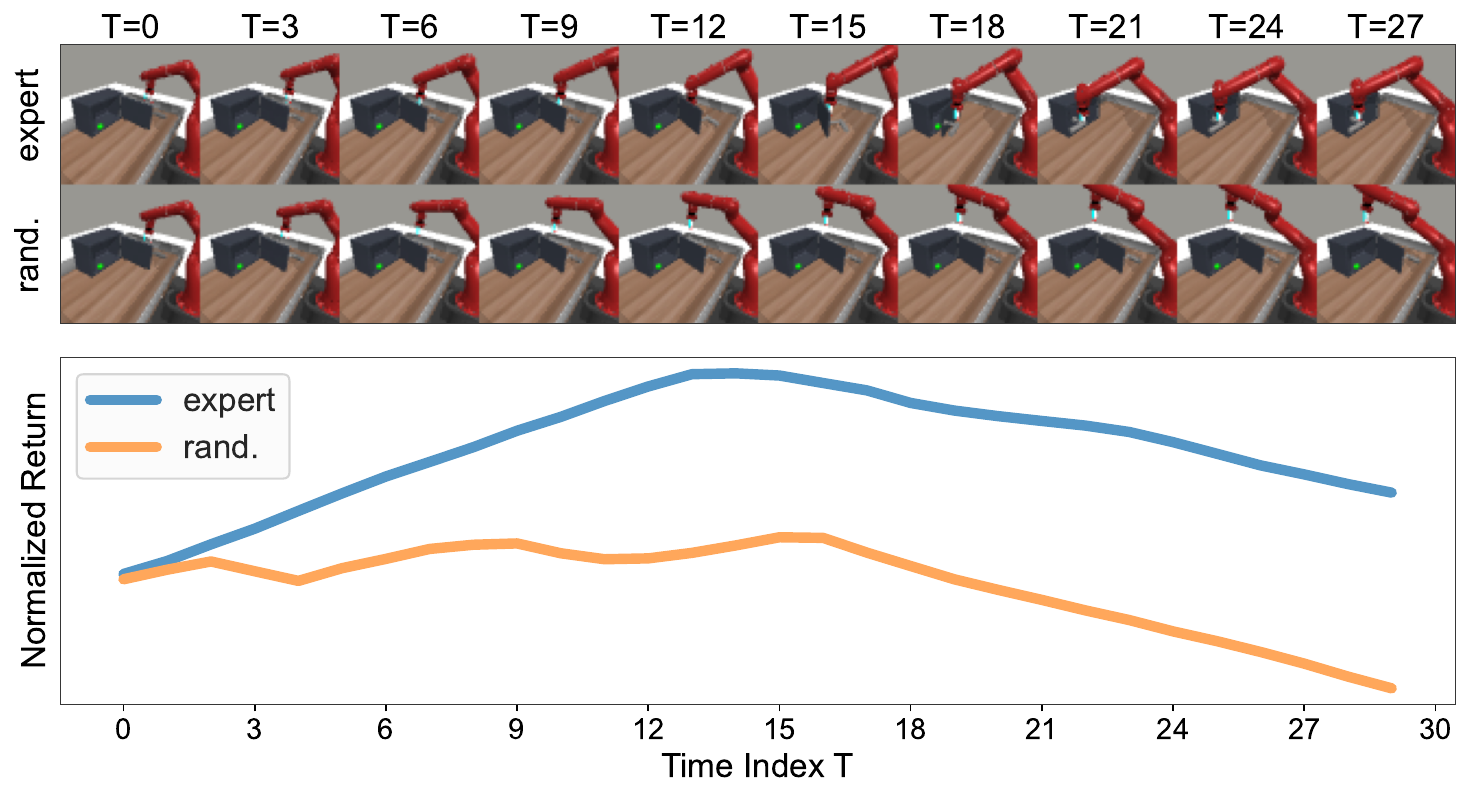}
     \end{subfigure}
     \\
     \begin{subfigure}[b]{0.496\linewidth}
         \centering
         \includegraphics[width=1.0\linewidth]{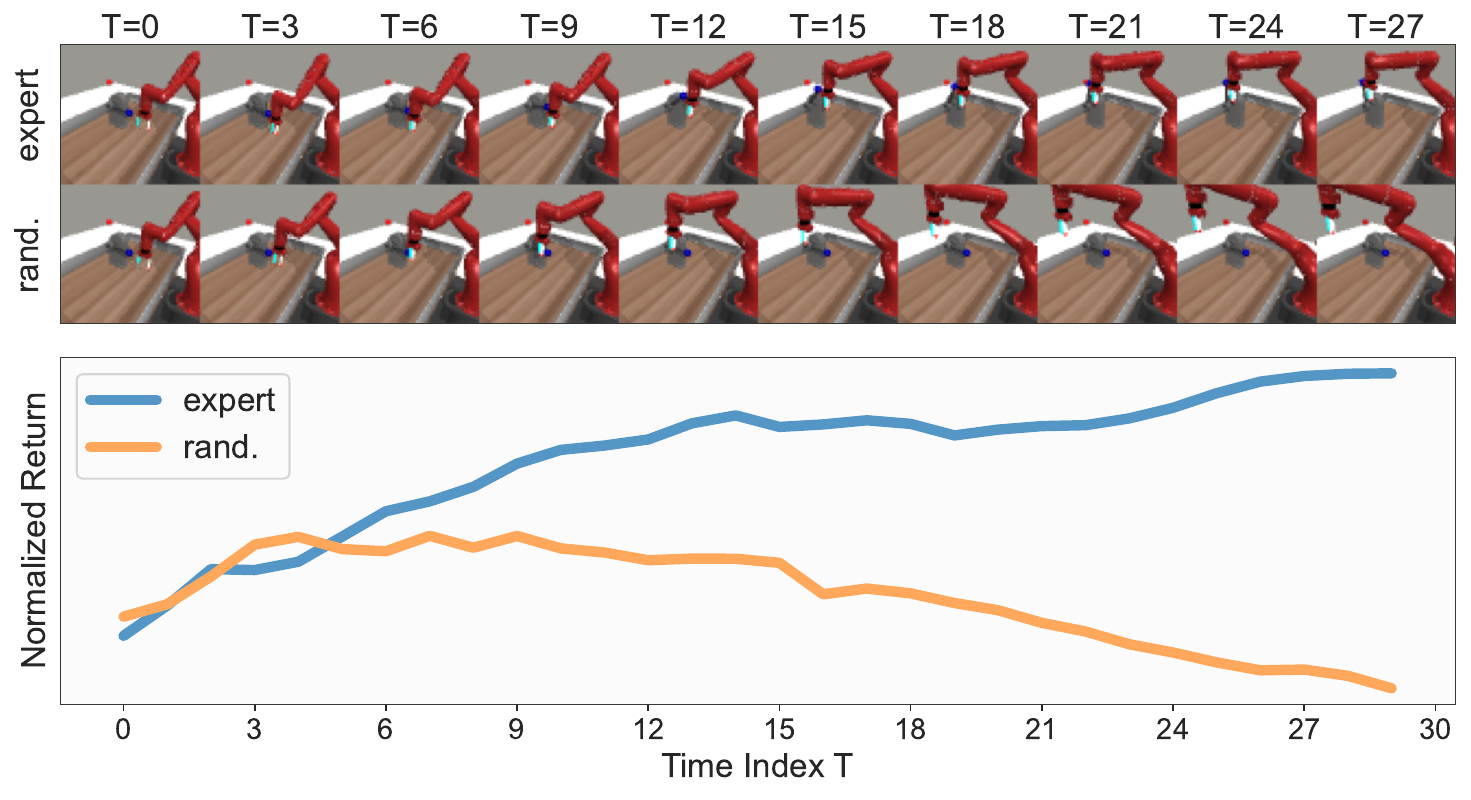}
     \end{subfigure}
     \begin{subfigure}[b]{0.496\linewidth}
         \centering
         \includegraphics[width=1.0\linewidth]{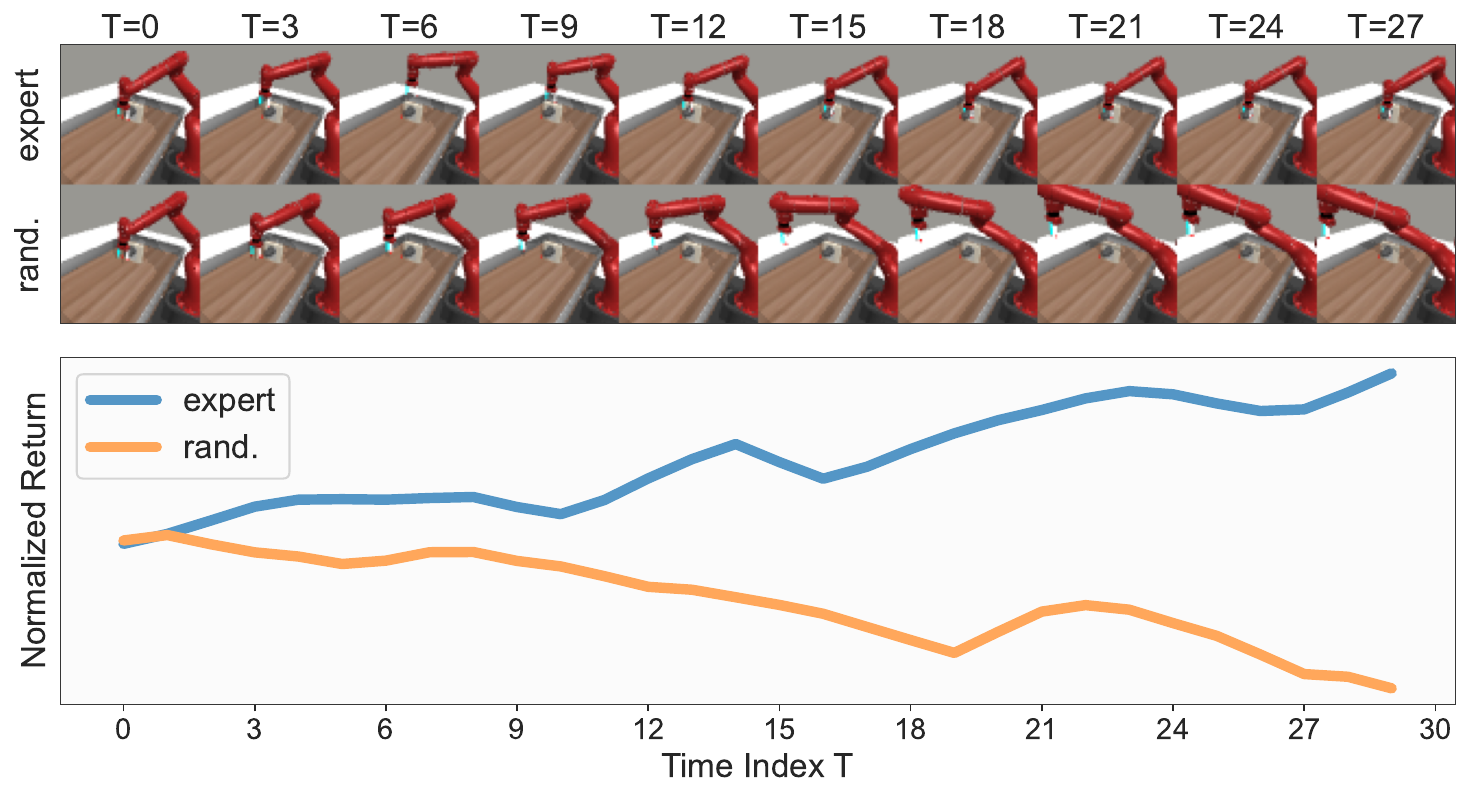}
     \end{subfigure}
     \\
     \begin{subfigure}[b]{0.496\linewidth}
         \centering
         \includegraphics[width=1.0\linewidth]{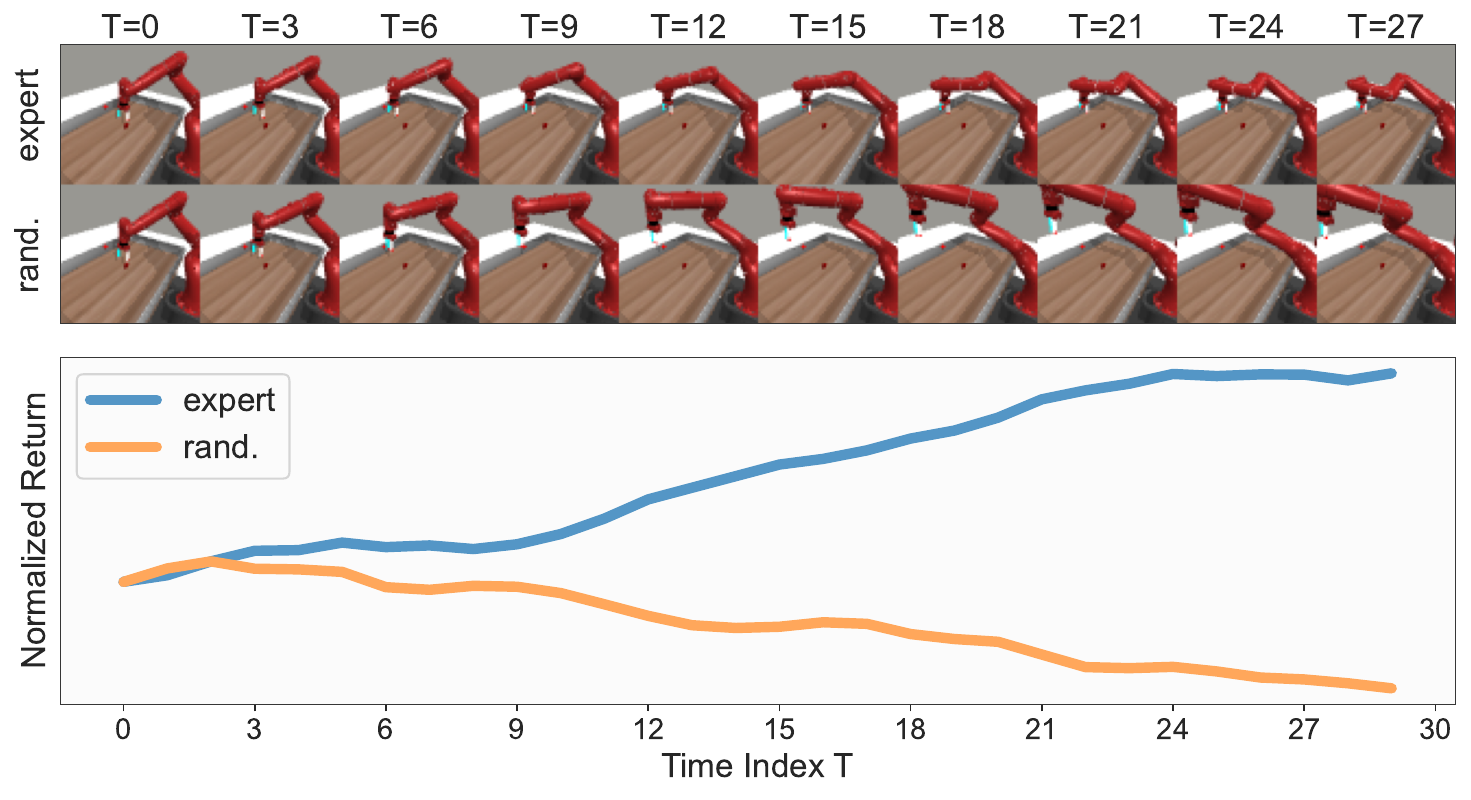}
     \end{subfigure}
     \begin{subfigure}[b]{0.496\linewidth}
         \centering
         \includegraphics[width=1.0\linewidth]{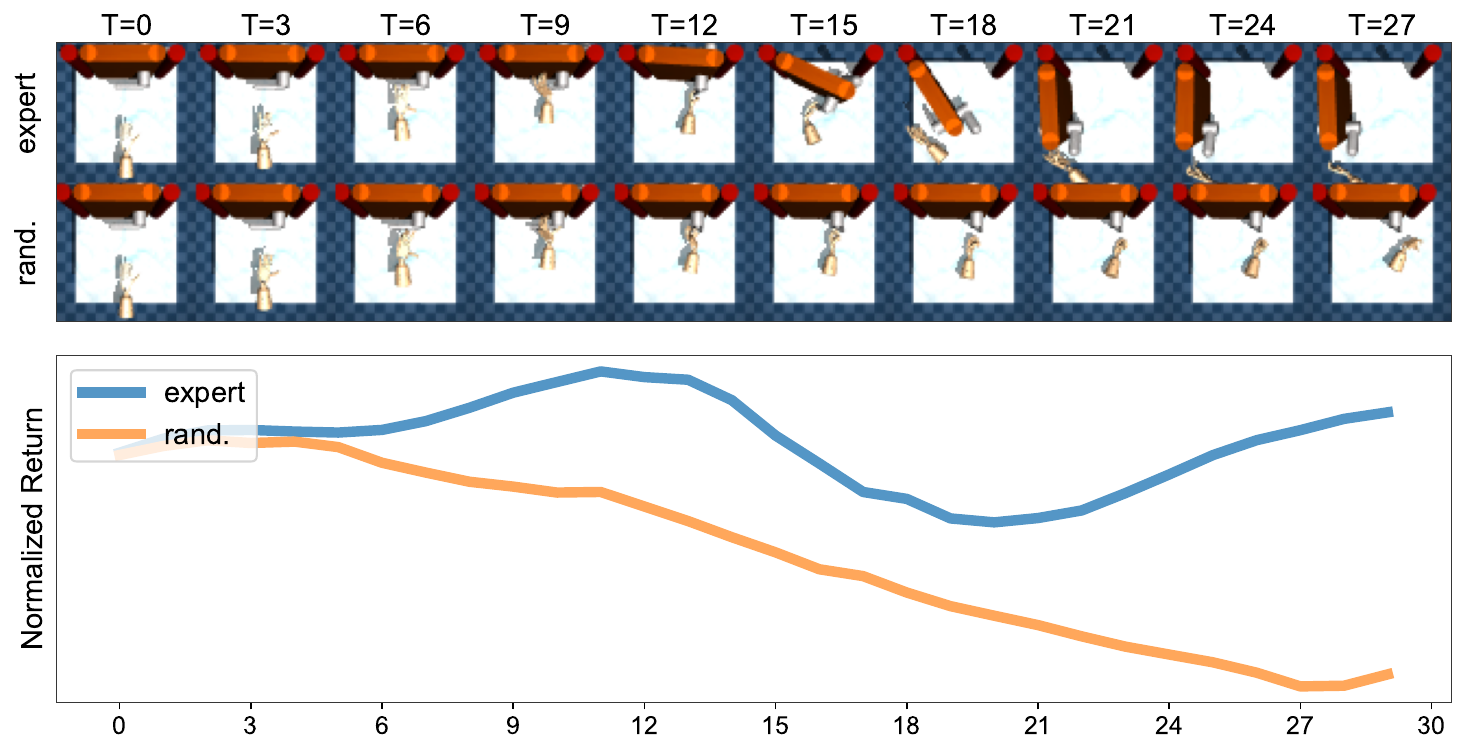}
     \end{subfigure}
     \\
     \begin{subfigure}[b]{0.496\linewidth}
         \centering
         \includegraphics[width=1.0\linewidth]{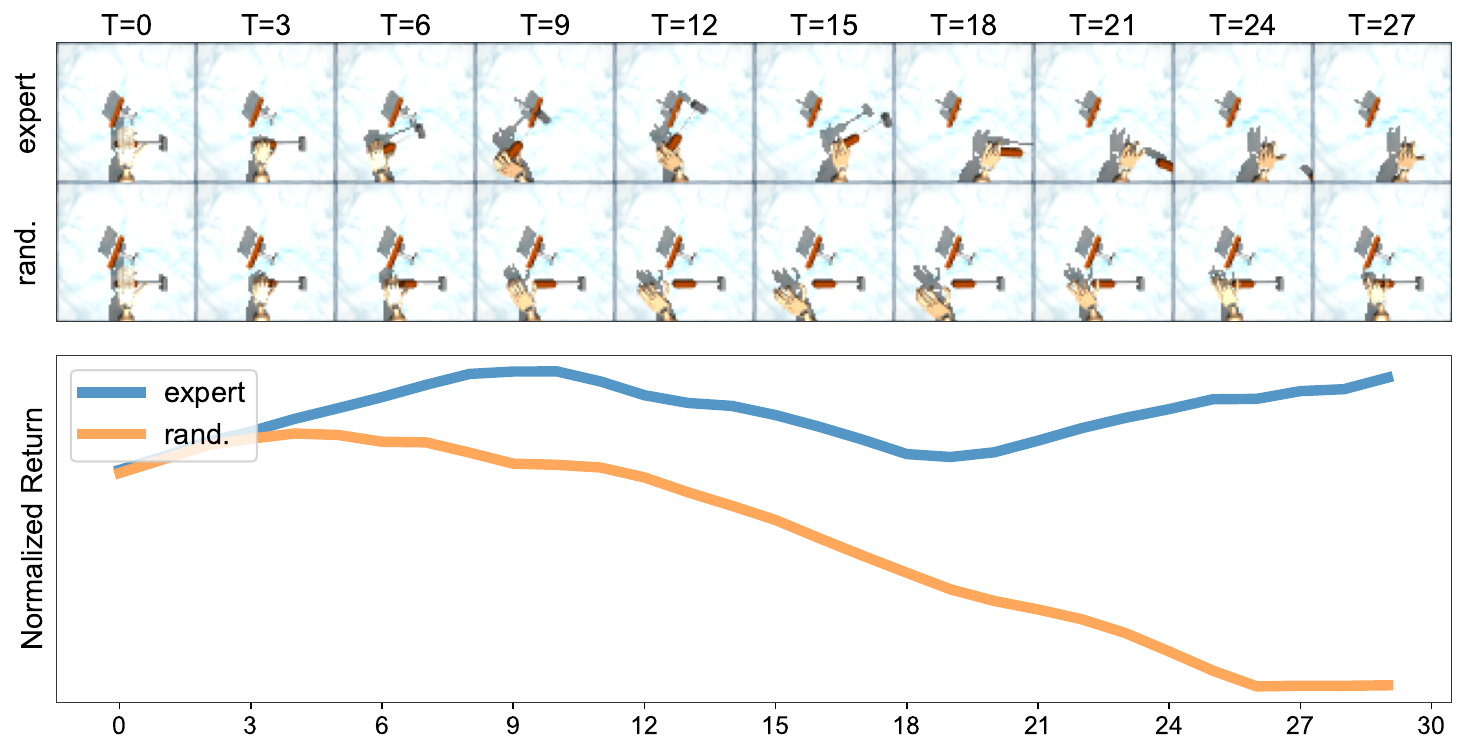}
     \end{subfigure}
     \begin{subfigure}[b]{0.496\linewidth}
         \centering
         \includegraphics[width=1.0\linewidth]{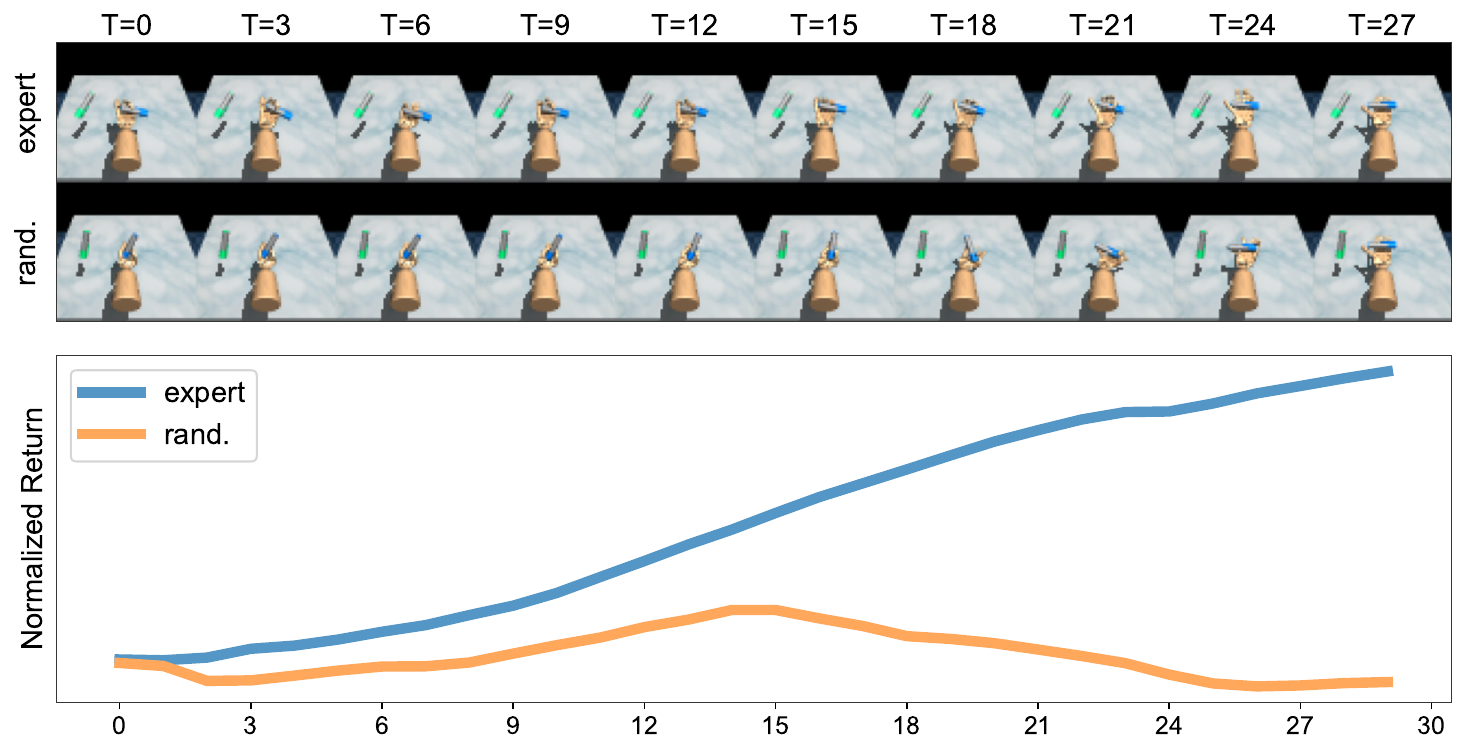}
     \end{subfigure}
 \caption{\textbf{Reward curve of 10 simulated tasks from MetaWorld and Adroit.} }
 \label{fig:reward-curve}
\end{figure*}

\paragraph{Simulation Results.} We first visualize the reward curve in 10 simulation tasks in Figure~\ref{fig:reward-curve}. Our proposed reward can greatly distinguish the expert-like and -unlike behaviors. Interestingly, we observe that two door-opening tasks show a return drop at the final execution stage. This may be attributed to the difficulty of modeling the dynamics of the door, suggesting that explicit modeling of environmental dynamics is worth investigating in the future.

\begin{figure*}[t]
     \centering
     \begin{subfigure}[b]{1\linewidth}
         \centering
         \includegraphics[width=1\linewidth]{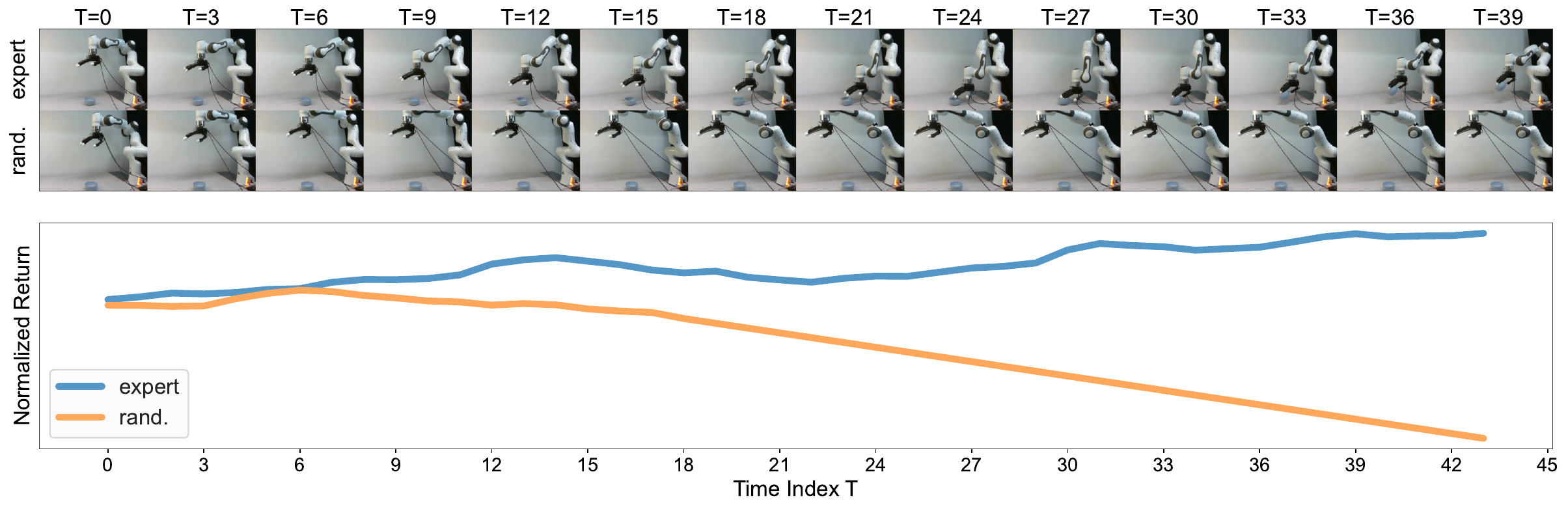}
     \end{subfigure}
     \centering
     \begin{subfigure}[b]{1\linewidth}
         \centering
         \includegraphics[width=1\linewidth]{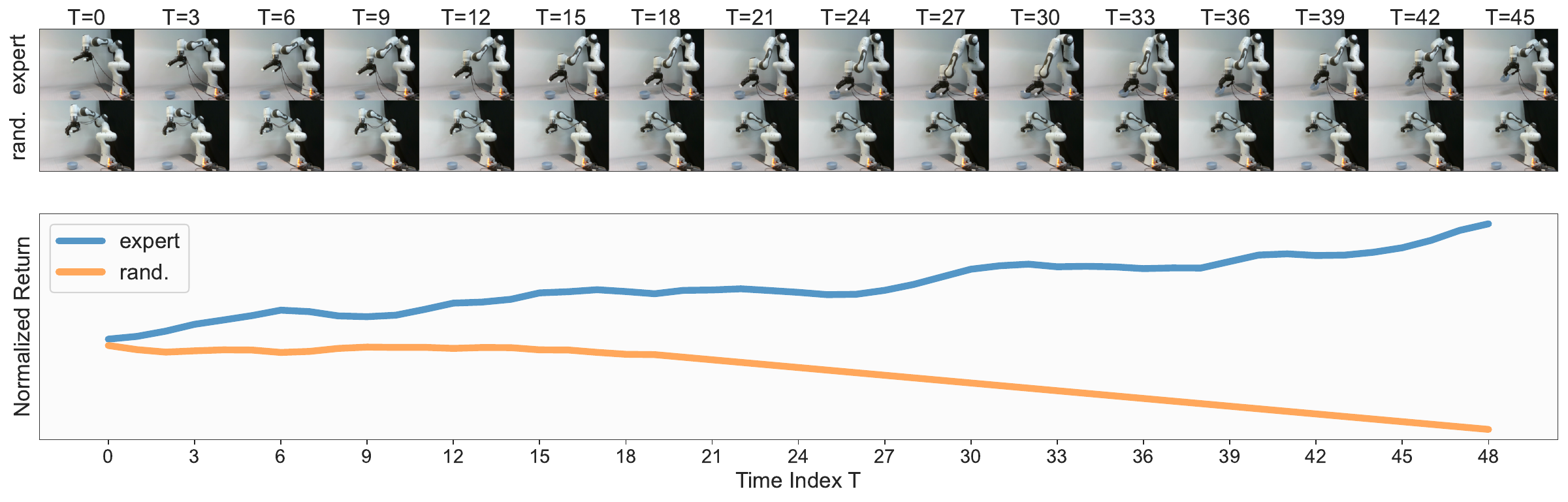}
     \end{subfigure}
     \centering
     \begin{subfigure}[b]{1\linewidth}
         \centering
         \includegraphics[width=1\linewidth]{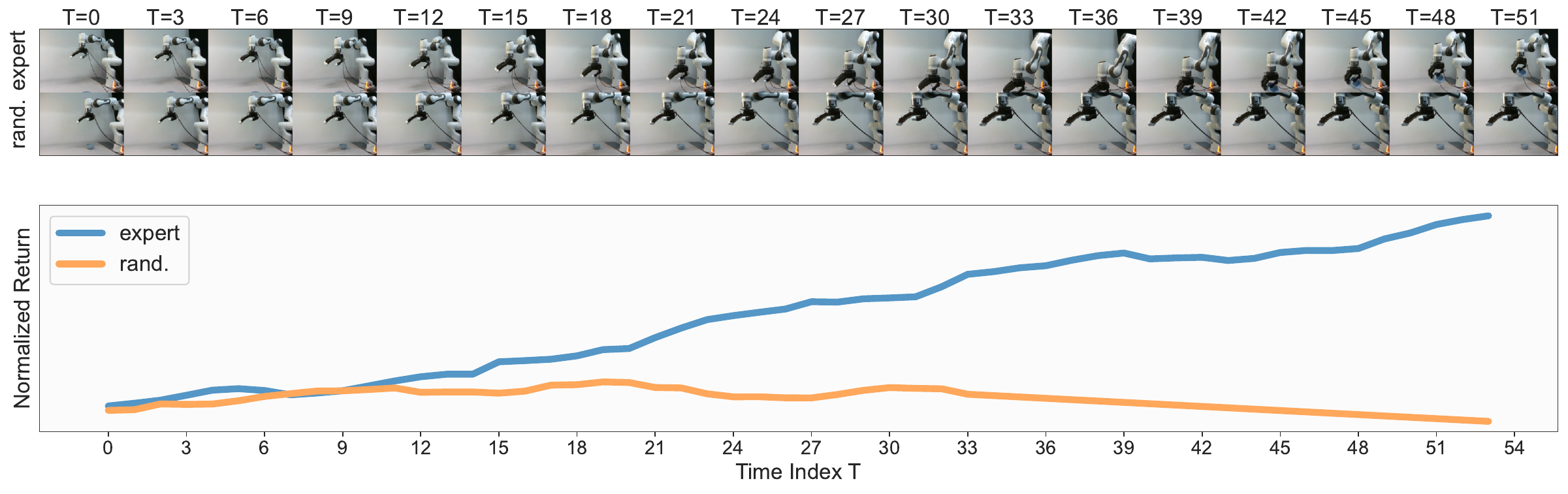}
     \end{subfigure}
 \caption{\textbf{Reward curve of real robot trajectories.} }
 \label{fig:reward-curve-real}
\end{figure*}

\pskip
\paragraph{Real Robot Results.} We collect 20 real robot video trajectories with an Allegro hand, a Franka arm, and a RealSense D435i camera. There is only one task, which is picking up a bowl on the table. 10 of the videos are success trajectories while the other 10 are random trajectories. We train our pipeline on the expert demonstrations and evaluate \ours on both expert and random trajectories. Visualization results in Figure~\ref{fig:reward-curve-real} show that \ours can correctly assign expert demonstrations relatively higher reward and random trajectories lower reward.

\begin{table}[h]
\caption{Hyperparameters for VQ-GAN.}
\centering
\label{table:vqgan}
\vskip -0.05in
\begin{tabular}{l|c}
\toprule 
\textbf{Hyperparameter} & \textbf{Value}  \\
\midrule
Input size  & $64\times 64 \times 3$  \\ 
Latent code size & $8\times 8$\\
$\beta$ (commitment loss coefficient) & $0.25$ \\
Codebook size & $1024$\\
Codebook dimension & $64$ \\
Base channels & $128$ \\
Ch. mult. & $[128, 128, 256, 256]$ \\
Num. residual blocks & $2$ \\
Use attention & True \\
Disc. start steps & $1000$ \\
Disc. loss weight & 0.1 \\
Reconstruction loss weight & $1$ \\
Perceptual loss weight  & $0.1$ \\
Training epochs & $200$ \\
Batch size & $32$ \\
Learning rate & $10^{-4}$ \\
Adam optimizer $(\beta_1, \beta_2)$ & $(0.5, 0.9)$ \\
\bottomrule
\end{tabular}
\end{table}

\begin{table}[t]
\caption{Hyperparameters for VQ-Diffusion.}
\label{table:vqdiffusion}
\vskip -0.05in
\centering
\begin{tabular}{l|c}
\toprule
\textbf{Hyperparameter} & \textbf{Value}  \\
\midrule
Num. transformer blocks & $16$ \\
Attention type  & Cross attention  \\
Num. attention head & $16$ \\
Embedding dimension & $128$ \\
Block Activation & GELU2 \\
Layer Normalization & Adaptive LN \\
Num. conditional frames & $2$ \\
Condition embedding dimension & $1024$ \\
Num. denoising steps & $10$ \\
Sampling noise type & Uniform \\
Adaptive auxiliary loss & True \\
Auxiliary loss weight & $10^{-3}$\\
Training epochs & $100$ \\
Batch size & $4$ \\
Learning rate & $4.5\times 10^{-4}$\\
AdamW optimizer $(\beta_1, \beta_2)$ & $(0.9, 0.96)$ \\
\bottomrule
\end{tabular}
\end{table}

\begin{table}[h]
\centering
\caption{Hyperparameters for DrQv2 with \ours.}
\label{table:drqv2}
\vskip -0.05in
\begin{tabular}{l|c}
\toprule
\textbf{Hyperparameter} & \textbf{Value}  \\
\midrule
\multicolumn{2}{l}{\underline{\textbf{Environment}}} \\
Action repeat & $3$ (MetaWorld) \\
                & $2$ (Adroit) \\
Frame stack & $1$ \\
Observation size & $64\times 64 \times 3$ \\
Reward type & Sparse \\
\multicolumn{2}{l}{\underline{\textbf{DrQv2}}} \\
Data Augmentation & $\pm 4$ RandomShift \\
Replay buffer capacity  &  $10^6$ \\
Discount $\gamma$ & $0.99$ \\
$n$-step returns & $3$ \\
Seed frames & $4000$ \\
Exploration steps & $2000$ \\
Feature dimension & $50$ \\
Hidden dimension & $1024$ \\
Exploration stddev. clip & $0.3$ \\
Exploration stddev. schedule & $\mathrm{Linear}(1.0,0.1,3\times 10^{6})$ \\
Soft update rate & $0.01$ \\
Optimizer & Adam \\
Batch size & $256$ \\
Update frequency & $2$ \\
Learning rate & $10^{-4}$ \\ 
\multicolumn{2}{l}{\underline{\textbf{RND}}} \\
CNN feature dimension & $7\times 7 \times 64$ \\
MLP size & $512 \times 512$ \\
Learning rate & $10^{-4}$ \\
\multicolumn{2}{l}{\underline{\textbf{\ours}}} \\
Reward coefficient $\alpha$ & $0$ (Pen) \\
                    &$ 0.95$ (Others)\\
Sampling noise & True (scale $1$) \\
Reward standardization $\bar{r}^{\mathrm{ce}}$ & True \\
Num. diffusion processss $M$ & $1$ \\

\bottomrule
\end{tabular}
\end{table}

